\documentclass[preprint,12pt]{elsarticle}

\usepackage[colorlinks=true,linkcolor=blue,urlcolor=blue,citecolor=blue]{hyperref}
\usepackage{amssymb}
\usepackage{amsmath}
\usepackage{cleveref}
\usepackage{color}
\usepackage{url}
\usepackage{tcolorbox}
\usepackage{tabularx}
\usepackage{colortbl}

\definecolor{green}{HTML}{44AA99}
\definecolor{yellow}{HTML}{DDCC77}
\definecolor{blue}{HTML}{88CCEE}
\definecolor{red}{HTML}{CC6677}
\definecolor{darkred}{HTML}{DC3220}

\usepackage{xspace}
\newcommand{\llama}{Llama-3\xspace}
\newcommand{\llamatext}{Llama-3.1-8B-Instruct\xspace}
\newcommand{\llamavision}{Llama-3.2-11B-Vision-Instruct\xspace}
\newcommand{\gpt}{GPT-4o\xspace}
\newcommand{\gptmini}{GPT-4o-mini\xspace}
\newcommand{\gemini}{Gemini 1.5\xspace}
\newcommand{\geminiflash}{Gemini 1.5 Flash\xspace}
\newcommand{\geminipro}{Gemini 1.5 Pro\xspace}
\newcommand{\llamaguard}{Llama-Guard-3\xspace}
\newcommand{\llamaguardtext}{Llama-Guard-3-8B\xspace}
\newcommand{\llamaguardvision}{Llama-Guard-3-11B-Vision\xspace}
\newcommand{\openaimod}{OpenAI moderation model\xspace}

\usepackage{array}
\usepackage{listings}
\usepackage[T1]{fontenc}
\usepackage{multirow}
\usepackage{mwe}
\usepackage[export]{adjustbox}
\usepackage{tabularray}

\journal{}

\begin{document}

\begin{frontmatter}

\title{Advancing Content Moderation: Evaluating Large Language Models for Detecting Sensitive Content Across Text, Images, and Videos}

\author[label1]{Nouar AlDahoul}\ead{naa9497@nyu.edu}
\author[label2]{Myles Joshua Toledo Tan}\ead{tan.m@ufl.edu}
\author[label2]{Harishwar Reddy Kasireddy}\ead{harishwarreddy.k@ufl.edu}
\author[label1]{Yasir Zaki\texorpdfstring{\corref{cor1}}{}}\ead{yasir.zaki@nyu.edu}

\cortext[cor1]{Corresponding author.}

\affiliation[label1]{organization=Computer Science Department, New York University Abu Dhabi,city=Abu Dhabi,country=UAE}

\affiliation[label2]{organization=Department of Electrical and Computer Engineering, Herbert Wertheim College of Engineering, University of Florida,
            city=Florida,
            country=USA}
            


\begin{abstract}
 The widespread dissemination of hate speech, harassment, harmful and sexual content, and violence across websites and media platforms presents substantial challenges and provokes widespread concern among different sectors of society. Governments, educators, and parents are often at odds with media platforms about how to regulate, control, and limit the spread of such content.
 Technologies for detecting and censoring the media contents are a key solution to addressing these challenges. Techniques from natural language processing and computer vision have been used widely to automatically identify and filter out sensitive content such as offensive languages, violence, nudity, sex, and addiction in both text, images, and videos, enabling platforms to enforce content policies at scale. However, existing methods still have limitations in achieving high detection accuracy with fewer false positives and false negatives. Therefore, more sophisticated algorithms for understanding the context of both text and image may open rooms for improvement in media content censorship to build a more efficient and accurate censorship system. In this paper, we evaluate existing large language model-based content moderation solutions such as \openaimod and \llamaguard and study their capabilities to detect sensitive contents.  Additionally, we explore recent large language models (LLMs) such as generative pre-trained transformer (GPT), Google Gemini, and Meta Llama in identifying inappropriate contents across media outlets. Various textual and visual datasets like X tweets, Amazon reviews, news articles, human photos, cartoons, sketches, and violence videos have been utilized for evaluation and comparison. The results demonstrate that LLMs outperform traditional techniques by achieving higher accuracy and lower false positive and false negative rates. This highlights the potential to integrate LLMs into websites, social media platforms, and video-sharing services for regulatory and content moderation purposes.
\end{abstract}

\begin{graphicalabstract}
\begin{figure}
    \centering
    \includegraphics[width=1\linewidth]{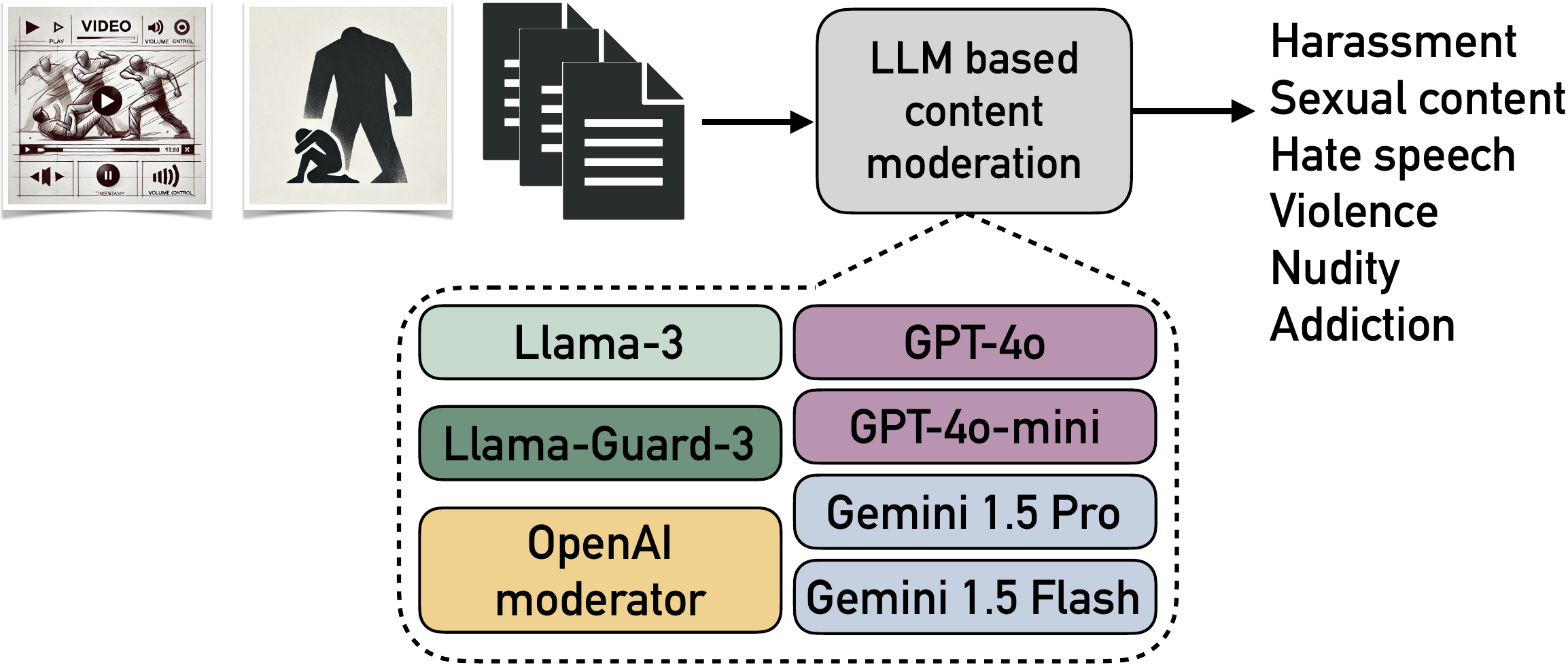}    
\end{figure}
\end{graphicalabstract}

\begin{highlights}
\item We uncovered the strength and weakness of using existing content moderation solutions, such as \openaimod and \llamaguard to censor various categories of text and images.
\item We explored the potential of using both open-sourced and proprietary large language models (LLMs) such as OpenAI \gpt, Google \gemini, and Meta \llama to censor and detect text that has inappropriate contents, including tweets, reviews, and articles. 
\item We brought to light the vision capabilities of large language models for censoring and detecting images and videos that have inappropriate contents, including adult and violent contents.
\item We evaluated the aforementioned detection and censorship solutions and compared them with state-of-the-art methods using various text and image datasets. The results show that LLMs have superior performance in censoring media contents containing text, images, and videos, outperforming existing content moderation solutions.
\end{highlights}


\end{frontmatter}

\section{Introduction}
The widespread dissemination of hate speech, harassment, harmful and sexual content, and violence across websites and media platforms presents substantial challenges and provokes widespread concern among different sectors of society~\cite{salter2013justice}. This type of content, readily available via the websites, social media, and other digital channels, affects individuals of all ages, with particularly profound and frequently negative impacts on children and adolescents~\cite{strasburger2010health}. The use of pornography among adolescents has been rising steadily, and the age at which they first encounter sexually explicit materials has been decreasing~\cite{greenfield2004inadvertent}. This issue is complex, intersecting with mental health concerns, societal norms, legal ramifications, and the technical hurdles associated with the surveillance and restriction of such materials~\cite{citron2018sexual}. First and foremost, the exposure of minors to sexually explicit, violent, and otherwise inappropriate content can detrimentally impact their psychological development and overall well-being~\cite{citron2018sexual}. Media violence exposure has been associated with heightened aggression, a numbing towards violence, and increased fear, leading to a skewed perception of reality and social norms~\cite{krahe2011desensitization}. From a legal perspective, the distribution and consumption of certain content types, especially child pornography, constitute criminal offenses across numerous jurisdictions~\cite{Child_Pornography}. The advent of the digital age has introduced complexities in detecting, monitoring, and prosecuting these offenses~\cite{powell2017sexual}. Law enforcement and legal systems face challenges in adapting to the swift technological advancements, resulting in a regulatory gap in online content oversight~\cite{Regulate_Tech}. 

On the technological front, identifying and filtering out explicit, violent, and sexual content involves intricate tasks that demand advanced algorithms and machine learning techniques~\cite{impact_of_algorithms}. The creation of this technology needs to strike a delicate balance between ensuring precision and effectiveness and upholding privacy rights and freedom of speech~\cite{Freedom_of_Speech}. The issues of false positives (erroneously identifying innocuous content as inappropriate) and false negatives (failing to recognize harmful content) pose significant challenges. The former may lead to undue censorship, while the latter risks exposing vulnerable populations to potentially harmful material. Despite these limitations, advancements in AI, such as improved image recognition models and more sophisticated algorithms for understanding human intent and context, hold promise for more efficient and accurate content moderation~\cite{OpenAI_Moderation,markov2023holistic,inan2023llama,dubey2024llama3herdmodels,aldahoulevaluation}. However, existing methods still have limitations in achieving high detection accuracy with fewer false
positives and false negatives. Therefore, this work opens rooms for improvement in media content censorship to build a more efficient and accurate censorship system.

In this study, our proposed content censorship solutions utilize state-of-the-art LLMs to identify inappropriate text data and the vision capability of these LLMs to recognize sensitive contents in the images and videos. Against this backdrop, we summarize our contributions as follows:

\begin{itemize}
    \item We explored LLM-based content moderation solutions such as \openaimod and \llamaguard.
    \item We unveiled the potential of using LLMs such as Google \gemini, \gpt, and \llama to detect inappropriate text such as tweets, reviews, and articles.
    \item We demonstrated a vision capability in LLMs and employed it in the task of visual content censorship, such as detection of images and videos containing nudity, pornography, violence, child abuse, alcohol, and drug abuse.
    \item We formulated a visual content censorship task as a visual question answering task using various LLMs.
    \item We utilized various textual and visual datasets containing text, images, and videos for evaluation and comparison.
\end{itemize}

To this end, we compile a visual and textual datasets including violence videos, images with nudity and pornography content, news articles from multiple news outlets, Amazon reviews, and X tweets, and examine violence, harm, hate speech, and sexual contents to answer the following research questions:

\begin{itemize}
\item RQ1: Are existing content moderation models such as \openaimod and \llamaguard sufficient to detect inappropriate textual and visual content in media?
\item RQ2: Are general-purpose LLMs such as \gemini, \gpt, and \llama able to overcome limitations available in the existing content moderation solutions?
\item RQ3: Do various types of LLMs exhibit consistent prediction percentages across different types of inappropriate content—such as hate speech, violence, and sexual content—using the same dataset?
\item RQ4: At what frequency do different categories of inappropriate contents, such as hate speech, violence, or sexual content, appear in news articles or product reviews?
\end{itemize}

This rest of the paper is organized as follows: In Section~\ref{sec:relatedwork}, we review previous works on content moderation methods. Section~\ref{sec:motivation} presents our research motivation. In Section~\ref{sec:methods}, we describe the datasets used to run the experiments. Section~\ref{sec:results} discusses the experimental results and compares the proposed solution with other baseline methods. Finally, conclusions and future works are discussed in Section~\ref{sec:conclusion}.

\section{Related Work}
\label{sec:relatedwork}

\subsection{Identifying inappropriate textual content}

The challenges of detecting and tracking hate speech domains, e.g., cyberbullying, abusive language, discrimination, sexism, extremism, and radicalization in text, are increasingly pressing for society, individuals, policymakers, and researchers. Several efforts have been made to employ automated techniques for detection and monitoring~\cite{jahan2023systematic}. Studies in literature have most widely used the SVM algorithm~\cite{jahan2023systematic,dinakar2012common,abozinadah2016improved,badjatiya2017deep} for classification. On the other hand, traditional feature representation methods such as TF-IDF ~\cite{dinakar2012common,davidson2017automated}, bag of words~\cite{pawar2018cyberbullying,ousidhoum2019multilingual}, and N-gram~\cite{alakrot2018towards,malmasi2018challenges} have been demonstrated widely. Additionally, different kinds of word embedding word2Vec~\cite{kamble2018hate,faris2020hate}, GloVe~\cite{badjatiya2017deep,rizos2019augment}, FastText~\cite{badjatiya2017deep,rizos2019augment}, and ELMO~\cite{zhou2020deep,dowlagar2021hasocone} with CNN~\cite{mulki2019hsab,yin2017comparative} and RNN~\cite{yin2017comparative,pitsilis2018effective} architectures have been explored in the literature.  Moreover, several works claimed BERT's~\cite{alatawi2021detecting,ranasinghe2019brums,dowlagar2021hasocone,polignano2019alberto} outperforming CNN and RNN models.

The increased connectivity has also facilitated the rapid dissemination of harmful and violence-inciting content. Several studies thoroughly investigate violence-inciting text using a diverse range of machine learning and deep learning models~\cite{das2023team,page2023mavericks,khan2024detection}.

Detecting content of profanity in speech or audio files for foul language censorship purposes is also an active research topic that has been explored using CNNs and RNNs~\cite{ba2021design,wazir2020spectrogram}.

The censorship of online sexual predatory behaviors and abusive language on social media platforms has become a critical area of research~\cite{nguyen2023fine,vogt2021early}. LSTM and BERT language models have been used for early sexual detection in chats and dialogue~\cite{vogt2021early,hamzah2021detection,yan2021bert}. 

The detection of hate speech, violence-inciting, and sexual contents in textual data has various impacts on media outlets such as tweets~\cite{ketsbaia2020detection,khezzar2023arhatedetector,al2022detection}, news articles~\cite{guelorget2021active,bello2020machine}, social media posts~\cite{del2017hate}, and messaging platforms~\cite{palazon2023identifying}.

Previously mentioned deep learning methods have outperformed traditional techniques in detecting hate speech, harmful, violence-inciting, and sexual contents in textual data. However, there is still room for improvement to enhance accuracy and reduce false positive and false negative rates. As such, this paper aims to improve textual content censorship using the large language models.

\subsection{Identifying inappropriate visual content}

The recognition of inappropriate visual content such as nudity and pornography in images and videos has seen significant advancements through various studies. This recognition is primarily categorized into four methods: color-based, shape-based, local feature-based, and machine learning-based. Color-based methods utilize pixel colors to identify skin regions, as discussed in several studies~\cite{ries2014survey,de2012statistical}. Shape-based methods are subdivided into techniques such as contour-based methods~\cite{arentz2004classifying,zaidan2015robust}, moments~\cite{wang1997system,zheng2006shape}, geometric constraints~\cite{fleck1996finding}, color segments~\cite{bosson2002non}, and MPEG7 features~\cite{kim2005detecting}, which each offer a unique approach to recognizing inappropriate content. Local feature-based methods leverage tools like the scale invariant feature transform (SIFT)~\cite{wijaya2015phonographic}, probabilistic Latent Semantic Analysis (pLSA)~\cite{lienhart2009filtering}, and bag of words (BoW) model~\cite{caetano2014pornography,caetano2016mid} to enhance detection accuracy. In recent years, machine learning-based techniques have become prevalent in the detection of adult content in images. Multiple-instance learning has been applied with notable success~\cite{jin2018pornographic,lipornographic}. 

Recent developments have seen a surge in deep learning-based solutions for pornography and nudity detection, employing convolutional neural networks (CNNs) and recurrent neural networks (RNNs) to enhance accuracy and efficiency~\cite{perez2017video,moustafa2015applying}. These approaches leverage the strengths of deep learning to interpret complex patterns in image and video content, significantly improving detection capabilities. The use of CNNs has been extensively researched, demonstrating their effectiveness in nudity detection~\cite{hor2021evaluation,aldahoul2019local,nian2016pornographic,aldahoul2020transfer,lyn2020convolutional,hor2022deep}. In addition,~\cite{aldahoul2021evaluation,aldahoul2021comparative,momo2023evaluation} has worked on the pornography detection in cartoon and sketch images. Moreover, an ensemble of CNNs has also been employed to improve detection rates~\cite{huang2016using}. Additionally, the incorporation of attention mechanisms has been explored to further refine the accuracy of nudity detection in images~\cite{wang2022pornographic}. These advancements highlight the dynamic nature of research in this field and the continuous efforts to improve content recognition technologies.

Violence detection is implemented across both surveillance and 
non surveillance environments, employing a spectrum of feature extraction methodologies that span traditional techniques to advanced deep learning approaches. Within traditional frameworks, methodologies such as scale-invariant feature transform (SIFT)~\cite{febin2020violence}, speeded-up robust features (SURF)~\cite{nadeem2019wvd}, and bag of words (BoW)~\cite{wang2012baseline} have been pivotal. Conversely, the deep learning paradigm leverages specialized recurrent neural networks (RNNs), notably long short-term memory networks (LSTMs)~\cite{patel2021real}, deep neural networks (DNNs)~\cite{ali2018violence}, Convolutional neural networks~\cite{bagga2022violence,cheng2021rwf}, and CNN-LSTM models~\cite{aldahoul2021convolutional,abdullah2023combination} to enhance the efficacy and accuracy of violence detection. Specifically, in the context of detecting bloody content, a tripartite feature extraction strategy is employed, encompassing static, motion, and audio features~\cite{wang2017review} to provide a comprehensive analysis. This multifaceted approach underscores the evolving landscape of violence detection methodologies, reflecting a transition from traditional techniques to sophisticated, deep learning-driven strategies for robust and effective violence detection. 

Previously mentioned deep learning methods have outperformed traditional techniques in detecting nudity, sexual contents, and violence. However, there is still room for improvement to enhance accuracy and reduce false positive and false negative rates. As such, this paper aims to improve visual content censorship in general and particularly nudity, sexual content, and violence detection using visual language models.

\subsection{Emerging role of LLMs in text and image processing}

Large Language Models (LLMs) have garnered significant attention for their impressive performance across various natural language tasks such as summarization~\cite{doss2024comparative}, classification~\cite{al2024analysis}, code generation~\cite{nejjar2023llms}, and data generation~\cite{huang2024generating}, particularly following the release of chatGPT in November 2022~\cite{minaee2024large}. Their ability to understand and generate language in a general-purpose manner is achieved by training billions of model parameters on vast amounts of text data.

LLMs have opened a new opportunity to address online sexual predatory chats and abusive texts by fine-tuning Llama2~\cite{nguyen2023fine}. They can also detect hate speech in text, according to a study~\cite{piot2024decoding}, which investigated  reactions of seven state-of-the-art LLMs (LLaMA 2, Vicuna, LLaMA 3, Mistral, GPT-3.5, GPT-4, and Gemini Pro). In this study, they also discussed strategies to mitigate hate speech generation by LLMs, particularly through fine-tuning and guideline guardrailing.

The vision capability in large language models refers to the extension of the transformer architecture, originally designed for natural language processing (NLP) tasks, to computer vision, enabling models to process and understand visual information in a manner similar to text~\cite{yenduri2024gpt}. While the original GPT models were focused on understanding and generating text~\cite{wu2023brief}, the concept of ``GPT for Vision''~\cite{deng2024vision,mu2024embodiedgpt,gupta2022towards} involves adapting this architecture to process and understand visual data. The key innovation in applying the GPT architecture to vision tasks lies in treating pixels or patches of an image as sequences, similar to how words or tokens are treated in NLP tasks. 

LLMs have also been integrated with traditional vision models in tasks like visual question answering (VQA)~\cite{antol2015vqa}, where models are trained to respond to questions based on image content. This highlights the synergy between LLMs and image recognition, requiring a deep understanding of both visual and textual data~\cite{antol2015vqa}.

Although there have been few attempts to detect hate speech and sexually explicit content using large language models, to the best of our knowledge, no prior works have specifically targeted the detection of inappropriate content in customer reviews and news articles using LLMs. Additionally, identifying nudity, violence, child abuse, and drug or alcohol abuse using large language models (LLMs) has not yet been explored. This presents a unique opportunity to expand the application of LLMs to these areas, addressing both textual and visual content across diverse contexts.

Integrating Large Language Models (LLMs), including Google \gemini~\cite{Gemini_15_technical_report,Introducing_Gemini_15}, \gpt~\cite{Hello_GPT-4o,GPT-4o}, and \llama~\cite{dubey2024llama3herdmodels} into censorship tasks has the potential to greatly enhance the performance and functionality of deep learning models. This paper explores recent state-of-the-art large language models to tackle complex challenges of natural language processing and computer vision, such as detecting inappropriate textual and visual contents, including violence, harassment, harm, sex, nudity, and hate speech. By leveraging the advanced understanding and generation capabilities of LLMs, the approach aims to improve textual and visual recognition in these domains.

\section {Research Motivation}
\label{sec:motivation}
The direct applications of LLMs in tackling such complex challenges of identifying inappropriate media contents such as tweets, posters, customers' reviews, articles, images, and videos are still an evolving area of research and practice. The current solutions provided by LLMs in various applications demonstrate, through their advanced comprehension and generative abilities, their potential to act as complementary—or even alternative—approaches to traditional and deep learning methods for censoring both textual and visual media. By harnessing the advanced language processing and contextual analysis capabilities of LLMs, researchers can significantly enhance the accuracy, efficiency, and flexibility of technologies employed in media content moderation and censorship.

\section{Materials and Methods}
\label{sec:methods}
This section describes the datasets used in the experiments conducted to evaluate the LLMs performance in textual and visual content censorship tasks. Moreover, the section discusses the baseline methods usually used in the literature and highlights our proposed LLM-based content censorship solutions. 

\subsection{Dataset Overview}

\subsubsection{Textual Datasets}

In this section, we discuss the textual datasets used to evaluate and explore LLMs for textual content censorship tasks. These datasets contain text from various sources, such as news articles, Amazon reviews, X tweets, articles' descriptions, and text that has inappropriate contents such as hate speech, violence, adult contents, harassment, and harmful contents. 

\begin{itemize}
    \item{Hate Speech and Offensive Language (HSOL) Dataset}\\HSOL is a public dataset designed for hate speech detection~\cite{davidson2017automated,hate_speech_offensive}. The authors started with a hate speech lexicon consisting of words and phrases identified by internet users as hate speech, sourced from Hatebase.org. Using the Twitter API, they retrieved tweets containing terms from the lexicon, resulting in a sample from 33,458 Twitter users. The timeline of each user was extracted, creating a collection of 85.4 million tweets. From this corpus, a random sample of 25,000 annotated tweets containing lexicon terms was selected and manually coded by CrowdFlower (CF) workers. The workers categorized each tweet into one of three categories: hate speech, offensive but not hate speech, or neither offensive nor hate speech. The dataset is divided as follows: hate speech 5.77\%, offensive language 77.43\%, and normal 16.80\%. In our study, the HSOL dataset was employed to evaluate the capabilities of numerous LLMs for hate speech content censorship and specifically in tweets usually available in social media. 

    \item{Gender-Based Violence Tweet Dataset}\\This is a public dataset proposed in a Kaggle challenge to address the gender-based violence (GBV) problem~\cite{Gender-Based_Violence}. The GBV remains a widespread issue globally, especially in developing and least-developed countries. The problem worsened in many regions during the COVID-19 pandemic. The goal of this GBV challenge is to develop a machine learning algorithm that categorizes 39,650 annotated tweets about GBV into five categories: sexual violence (82.34\%), physical violence (15\%), emotional violence (1.64\%), economic violence (0.55\%), and harmful traditional practices (0.47\%). In this work, the GBV dataset was utilized to evaluate the capabilities of numerous LLMs for violence content censorship and specifically in tweets usually available in social media. 

    \item{Adult Content Dataset}\\We used a dataset of 850 articles' descriptions available on a Hugging Face platform and annotated with two different categories, namely: non-adult (512 samples) and adult (338 samples)~\cite{Adult-content-dataset}. This dataset was utilized in this work to evaluate the capabilities of LLMs for adult content censorship and specifically in the description of articles that are usually available on the websites. 

    \item{Amazon Product Reviews Dataset}\\This is a large-scale Amazon Reviews dataset collected in 2023. This dataset contains 48.19 million items and 571.54 million reviews from 54.51 million users~\cite{hou2024bridging}. The timespan is from May 1996 to September 2023. We selected a subset of this dataset to focus on reviews related to beauty products, with around 700,000 various reviews. This dataset is not annotated regarding content censorship. We aim to show the capabilities of LLMs in customers' review censorship tasks by exploring if the reviews have inappropriate contents such as hate speech, violence, or sexual contents or if they have already been filtered by the Amazon content moderation system. Amazon offers pre-trained models through an API interface for image and video moderation~\cite{Amazon_REcognizer_Content_Moderation}. Amazon Recognizer Content Moderation is able to automatically detect harmful images in product reviews with higher accuracy.  Additionally, Amazon Transcribe Toxicity Detection is a technique used by Amazon to identify toxic content and categorize it into seven areas: sexual harassment, hate speech, threats, abuse, profanity, insults, and graphic material~\cite{Amazon_Transcribe_Toxicity_Detection}.

    \item{News Articles from various Media Agencies}\\We scraped news articles from websites of five news outlets having various political leanings, namely The Daily Beast (Left), CCN (Lean left), Newsweek (Central), The Washington Times (Lean right), and Fox News (Right). The articles selected from the websites are under the general US category, which has a subcategory of crime. We collected the articles from 2013 to 2023. The numbers of collected articles are  30,600, 16,750, 14,400, 57,650, and 103,550 for CNN, Fox News, The Daily Beast, Newsweek, and The Washington Times, respectively. This dataset is not annotated regarding content censorship. We aim to show the capabilities of LLMs in long article censorship tasks by exploring if the news articles have inappropriate contents such as hate speech, violence, or sexual contents. 
\end{itemize}

\subsubsection{Visual Datasets}
In this section, we describe the visual datasets utilized in this work to evaluate and explore LLMs for visual content censorship tasks. These datasets contain images that have inappropriate contents such as nudity, pornography, violence, graphic violence, drug, alcohol, and child abuse contents. 

\begin{itemize}
    \item{Human Photo Dataset}\\We utilized a public dataset of traditional photos of naked humans meticulously gathered from the internet to encompass diverse backgrounds, such as forests, snowy landscapes, beaches, supermarkets, indoor settings, and streets~\cite{aldahoul2020transfer}. This dataset is considered challenging due to variations in numerous factors, including human positions (standing, sitting, lying, etc.), the proportion of nude individuals relative to the frame size, clothing, skin tones, and backgrounds. The dataset includes a total of 800 annotated images: 400 normal images and 400 nude images. We made a comparison between various LLMs from one side and with state-of-the-art methods in terms of accuracy, F1 score, false positive rate, and false negative rate. 

    \item{Cartoon Dataset}\\A cartoon dataset of 900 cartoon images~\cite{aldahoul2021evaluation} proposed for nudity and pornography detection was utilized in this work for the visual content censorship task. The images scraped from the web were annotated and divided into two sets: 400 pornographic and 500 normal. The dataset includes images with various skin colors, persons' positions, number of persons, scale of nude regions, and levels of lightening. 

    \item{Sketch Dataset}\\We utilized a sketch dataset that includes sketch images of people in various positions and environments with two classes, namely normal and pornographic~\cite{momo2023evaluation}. It consists of 1000 annotated images that were then divided into two sets: 500 pornographic and 500 normal. 
\end{itemize}

\noindent Samples of the aforementioned datasets: human, cartoon, and sketch can be seen under: \url{https://github.com/comnetsAD/LLM-Content-Moderation}. The black and red rectangles on the naked regions were added manually for public consideration.


\begin{itemize}
    \item{Violence Dataset}\\In this research, two public datasets, RWF-2000~\cite{cheng2021rwf} and Real-Life Violence Situations (RLVS)~\cite{soliman2019violence}, were combined to create a comprehensive annotated violence dataset comprising 4,000 videos: 2,000 violence and 2,000 non-violence samples. RWF-2000~\cite{cheng2021rwf} is a large-scale video dataset with 2,000 videos captured by surveillance cameras in real-world settings. On the other hand, Real-Life Violence Situations (RLVS)~\cite{soliman2019violence} is a dataset containing 2,000 videos sourced from YouTube, consisting of 1,000 videos of street fights and other violent scenarios and 1,000 videos of non-violent activities like sports, eating, and walking. In this work, we used a testing dataset, which is a subset of RWF-2000 and RLVS with 800 videos divided into 400 violence and 400 normal. 

    \item{Graphic Violence Dataset}\\We compiled this dataset by scraping 350 images of bloody content (scenes associated with bleeding, injuries, violent or graphic scenes, crime scenes) from the web to assess the effectiveness of LLMs in content censorship, focusing on detecting the graphic violence in the images and videos. Samples of videos from the violence fighting videos in the first row and graphic violence images in the second row can be seen under: \url{https://github.com/comnetsAD/LLM-Content-Moderation}. 


    \item{Alcohol, Drug, and Child Abuse Dataset}\\We created this dataset by scraping 423 images from the web to evaluate various LLMs for content censorship and to detect the inappropriate contents related to abuse in the images and videos. The images in this dataset have three types of abuse contents: child, drug, and alcohol. Samples of images from each abuse type can be seen under: \url{https://github.com/comnetsAD/LLM-Content-Moderation}. The first, second, and third rows depict alcohol, child abuse, and drug abuse, respectively.


\end{itemize}

\subsection{Methods}

This paper examines harmful content across both text and visual modalities, offering a thorough foundation for identifying the limitations of existing content censorship systems. It addresses these gaps by exploring the potential of LLM-based content moderation, which enhances detection accuracy and minimizes false positive and negative rates through the integration of vision-language contexts.
 
\begin{figure}[!htb]
    \centering
    \includegraphics[width=1\linewidth]{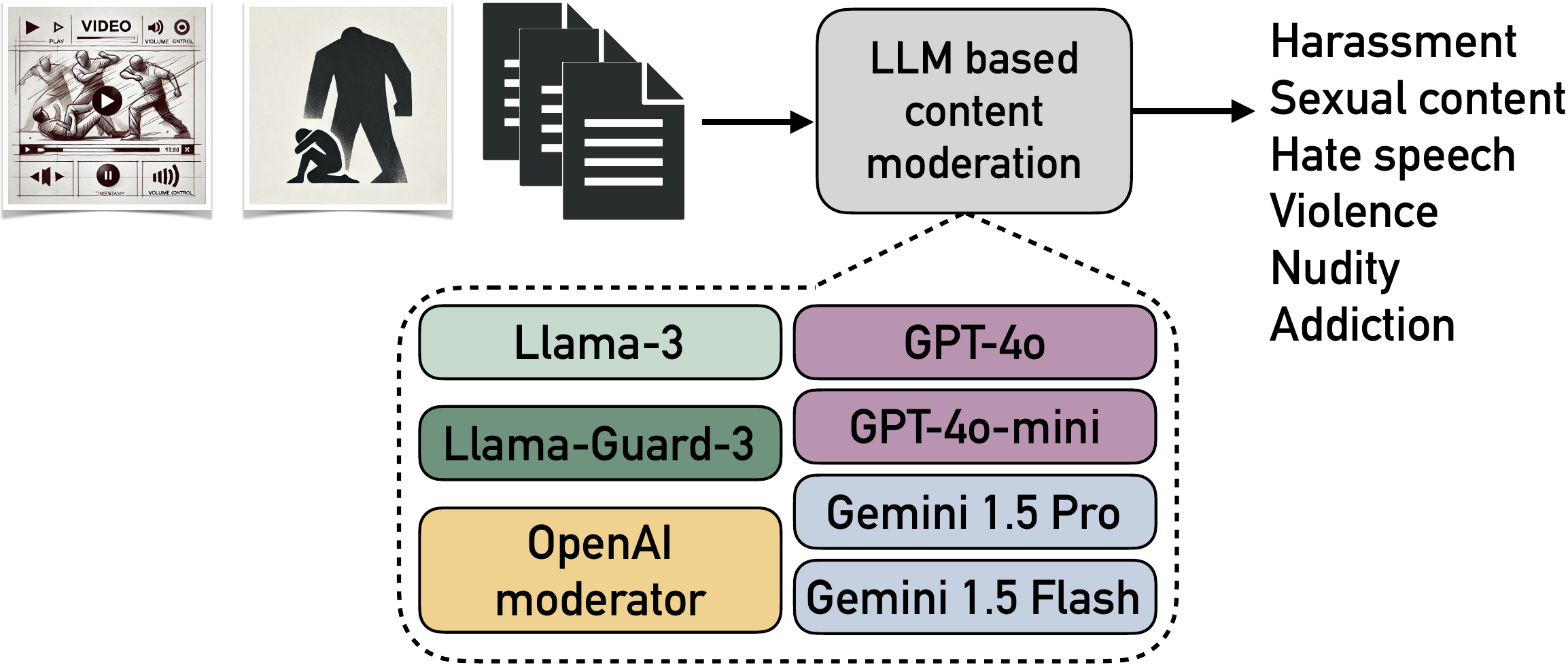}
    \caption{The Block Diagram of our Proposed Solution}
    \label{fig:solution}
\end{figure}

The proposed solution for media censorship is an AI system that combines language and visual processing to improve content understanding and generation. This system is designed to detect inappropriate content in all types of media, such as text, images, and videos, based on provided prompts. We employed large-language models to harness their natural language processing abilities, enabling contextual interpretation and analysis within images. Figure~\ref{fig:solution} presents the block diagram illustrating the proposed solution for textual and visual content censorship.

We utilized the models covered in this section  to assess the performance of large-language models in detecting inappropriate textual and visual content, which is central to this research.

As shown in Figure~\ref{fig:solution}, the text, images, and videos required censoring are applied to the inputs of each LLM, namely OpenAI Omni moderation~\cite{OpenAI_Moderation}, Meta \llamaguard~\cite{Llama_Guard,dubey2024llama3herdmodels}, OpenAI \gpt~\cite{GPT-4o}, Google \gemini~\cite{Gemini_15_technical_report}, and \llama~\cite{dubey2024llama3herdmodels}. The prompt is also applied to the inputs to drive the entire process. We evaluated each of these LLMs separately and compared their outcomes against the ground truth. These LLMs represent the well-known LLMs available in the literature.

\subsubsection{\openaimod with Textual and Visual Date}

OpenAI's moderation system found by OpenAI serves as a foundational framework, leveraging machine learning models designed to classify inputs (text and/or image) if potentially harmful across several categories~\cite{OpenAI_Moderation}. ``Omni-moderation-latest'' model is the latest OpenAI moderation model that supports more categorization options and multi-modal inputs~\cite{OpenAI_Moderation}. The categories contain harassment, harassment/threatening, hate, hate/threatening, illicit, illicit/violent, self-harm, self-harm/intent, self-harm/instructions, sexual, sexual/minor, violence, and violence/graphic~\cite{OpenAI_Moderation}. 

In this study, we utilized OpenAI's content moderation tools and particularly the ``Omni-moderation-latest'' model to explore and evaluate the LLM-based content moderation in its capability to detect inappropriate content, such as sexuality, violence, hate speech, and other harmful materials~\cite{OpenAI_Moderation}. 

First for text moderation, we employed the ``omni-moderation-latest'' model~\cite{OpenAI_Moderation} to automatically evaluate text input such as articles, Amazon customers' reviews, and X tweets by assigning a confidence score based on the likelihood that the content fits into one of several harmful categories, including hate speech, adult contents, and violence. These confidence scores are used to flag content that may need intervention or filtering~\cite{OpenAI_Moderation}.

Second, for visual content moderation, the ``omni-moderation-latest'' model analyzes visual content such as images and videos, detecting harmful content like nudity, pornography, violence, child abuse, and drug and alcohol abuse. ``Omni-moderation-latest'' model incorporates object and scene recognition to classify images based on the same harmful content categories used in text moderation~\cite{OpenAI_Moderation}.

\subsubsection{\llamaguard with Textual and Visual Date}

\llamaguard is an LLM-based safeguard model developed by Meta and available on Hugging Face. It is designed to handle multi-modal tasks by combining natural language processing with visual recognition capabilities~\cite{inan2023llama,dubey2024llama3herdmodels}. \llamaguard is also designed for human-AI conversational scenarios. The model features a safety risk taxonomy, which serves as an effective tool for categorizing specific safety risks identified in LLM prompts, enabling prompt classification~\cite{inan2023llama}.

\llamaguard model uses multi-modal transformers to interpret visual and textual inputs concurrently~\cite{inan2023llama}, making them well-suited for our study's focus on detecting inappropriate content across text and image modalities. 

\llamaguardtext, with 8 billion parameters, exhibited strong performance in language tasks~\cite{Llama_Guard}. It is particularly useful in scenarios where computational resources are limited while still maintaining effective moderation capabilities. 

On the other hand, the \llamaguardvision is a larger version with enhanced capabilities in both text and image recognition~\cite{Llama_Guard_vision}. This model leverages its 11 billion parameters to provide more accurate and robust classification of inappropriate content in images and associated textual descriptions. The model was specifically tested for its ability to detect harmful imagery, such as explicit scenes or depictions of violence, and align these detections with accompanying textual cues for a comprehensive understanding of content.

In this study, we tested the \llamaguardtext~\cite{Llama_Guard} and \llamaguardvision~\cite{Llama_Guard_vision} as a part of our effort to detect inappropriate textual and visual content, such as hate speech, sexual, and violent contents.

\subsubsection{Llama Instruct}

Llama 3.1 is an auto-regressive language model, developed by Meta, built on an optimized transformer architecture~\cite{Llama}. It consists of multilingual LLMs that include pre-trained and instruction-tuned generative models, designed for text input and text output. \llamatext is a fine-tuned version that employs supervised fine-tuning (SFT) and reinforcement learning with human feedback (RLHF) to better align the model with human preferences for helpfulness and safety~\cite{Llama}.

\llamavision is built upon the Llama 3.1 text-only model~\cite{Llama_vision} and comprises multi-modal generative models, capable of processing both text and images as input and generating text output. \llamavision is an instruction-tuned model, specifically optimized for tasks such as visual recognition, image reasoning, captioning, and answering general questions about images~\cite{Llama_vision}. For image recognition tasks, \llamavision incorporates a separately trained vision adapter that integrates with the pre-trained Llama 3.1 language model.

In this study, we evaluated both \llamatext~\cite{Llama} model and \llamavision~\cite{Llama_vision} to complement our effort to detect inappropriate content, such as sexuality, violence, and hate speech, within both text- and vision-based data. The model is able to ensure that harmful content is filtered out based on a combination of object recognition in images and semantic analysis in text.

\subsubsection{\gemini}

The \gemini model, developed by Google, integrates advanced vision and natural language processing techniques, making it particularly effective in understanding both text and images~\cite{Gemini_15_technical_report}. It also incorporates Google's advancements in instruction-following tasks, ensuring that specific content moderation guidelines were adhered to when evaluating both text and image inputs~\cite{Gemini_Vertex}. 

With its sophisticated architecture, \geminipro excels in scenarios requiring high-level contextual understanding of images in conjunction with text. On the other hand, \geminiflash is a lightweight, smaller, and faster model, optimized for speed and efficiency~\cite{Gemini_15_technical_report}. 

The Gemini API provides safety settings that can be adjusted. It has several levels of blocking content, such as no block, only high probability of unsafe content, only medium or high probability, all low, medium, or high probability~\cite{safety-settings}. The Gemini API's safety filters cover the following categories: harassment, hate speech, sexual explicit, and dangerous content.  

In this study, we further evaluated \geminipro and \geminiflash to assess their capabilities in detecting inappropriate textual and visual contents that are usually available in media outlets.

\subsubsection{\gpt}

\gpt is an OpenAI's advanced language model builds upon the robust architecture of earlier GPT models, expanding its proficiency in both text and vision-language tasks~\cite{GPT-4o}. \gpt employs an extensive parameter count and deep learning techniques that enable it to recognize text and visual content based on complex contextual understanding. Its vision-language fusion allows it to interpret both image-based and textual data concurrently. Leveraging \gpt's instruction-following capabilities, the model was fine-tuned for the specific tasks, ensuring adherence to predefined guidelines~\cite{GPT-4o}.

In this study, we evaluated \gpt~\cite{GPT-4o,Hello_GPT-4o}, specifically in the context of its multi-modal capabilities for detecting inappropriate textual and visual contents such as adult content, violence, and hate speech. This was particularly beneficial in scenarios where inappropriate visual elements were paired with text.

\section{Results and Discussion}
\label{sec:results}
This section presents the evaluation of several LLMs on tasks related to textual and visual content censorship. It also demonstrates the comparison results in terms of their accuracy, recall, precision, F1 score, false positive rate (FPR), which is a proportion of negative instances incorrectly classified as positive, and false negative rate (FNR), which is a proportion of positive instances incorrectly classified as negative. 

State-of-the-art LLMs such as \llamaguard,  \openaimod, \gpt, \gemini, and \llama are utilized and compared to tackle the above challenging censorship tasks. Additionally, we compared the LLM-based solutions with existing state-of-the-art methods. 

Both \gemini and \openaimod can detect multiple categories for the same instance. While \openaimod provides a score for each detected category, \gemini assigns a Low, Medium, or High level for the detected category. Other LLMs used in our experiments generate only one category, as determined in the prompt given.

Since \gemini lacks specific categories for `harm', `violence', and `graphic violence', these categories are grouped under the broader ``dangerous'' category.   

We carried out multiple experiments to assess the capabilities of various LLMs for textual and visual content censorship, with a specific focus on identifying hate speech, sexual, and violent contents. Formulating the censorship application as a visual question-answer task allows to leverage the  LLMs with their capabilities of understanding and processing both the image and associated text.  

\subsection{Visual Content Analysis}

First, we carried out several experiments to demonstrate the potential of using LLMs, such as \openaimod, \llamaguard, \gpt, \gemini, and \llama for visual content censorship and particularly nudity, pornography, violence, graphic violence, and abuse. We analyzed the results as shown in~\Cref{tab:visual_violence_LLMs,tab:visual_human_LLMs,tab:visual_cartoon_LLMs,tab:visual_sketch_LLMs,tab:visual_abuse_LLMs}.

For \openaimod, which is a pre-trained GPT for moderation task, the API is designed without the need to feed the LLM by any prompt because the LLM was already trained for moderation purposes. For the visual violence task, the category generated by \openaimod is `violence'. On the other hand, the nudity/pornography task, \openaimod generated a 'sexual' category.

For \llamaguard, applying empty prompt results in bad performance for image moderation. As a result, we used the simple prompt ``describe the image'' in \llamaguard to use its vision feature to analyze images, which is the main goal of detecting inappropriate visual contents. \llamaguard is also a pre-trained for moderation task, and it generated a limited, pre-defined set of categories as a response to prompts, chats, and images fed to its input. For  the nudity/pornography task, it generates `unsafe with S12' which refers to sexual content. However, it generates `unsafe with S8' for several images; this indicates its detection of intellectual property issues with images fed to it. 

For \gemini visual violence task, we set safety settings to be `BLOCK-LOW-AND-ABOVE'. In this case, GEMINI will block prompts or images when the probability score is either low, medium, or high.  As a result, a few videos are assigned to the `normal' category mentioned in the prompt. Conversely, most videos were blocked, even the ones with low probability scores. The videos blocked with low probability scores are considered to have normal content, while others blocked with medium or high probability scores are categorized as having violent content. On the other hand, for \gemini visual nudity/pornography task, we set safety settings to be `BLOCK-NONE' to prevent \gemini from blocking any prompt and force it to select from the list defined in the given prompt. However, because \geminipro continues rejecting to answer for a set of images, we changed the prompt for unanswered samples to be `Does this image have NSFW content? Answer by Yes or No'. NSFW stands for Not Suitable For Work. In this case, \geminipro was able to answer and find the categories for all images given.

\subsubsection{Detection of Violence in Videos}

The first experiment aims to study the LLMs capability to detect the violence in the videos. We used 800 testing videos from RWF-2000~\cite{cheng2021rwf} and Real-Life Violence Situations (RLVS)~\cite{soliman2019violence}. 

We selected five LLMs, including large and small-size LLMs such as \gpt, \gptmini, \geminipro, \geminiflash, \llama that support vision in general and video particularly, by considering multiple images at input driven by a textual prompt. The video is divided into frames, taking one frame only from each ten frames. The frames are stored and fed to the input of LLMs for each inference. We did not use \openaimod, and \llamaguard in this experiment because they do not support multiple images as input, which addresses \textbf{RQ1}.

The prompt used for violence detection in the videos is:

\begin{tcolorbox}[colback=orange!5!white, colframe=orange!75!black, title={violence prompt}, rounded corners, boxrule=1pt, boxsep=1pt]
Find the category of these video's frames based on the following list: [`violence',`fighting',`normal'], answer only using a single word from this list.
\end{tcolorbox}

Table~\ref{tab:visual_violence_LLMs} shows the results of comparison between various LLMs. The objective of a robust content censorship system is to reduce and balance both FPR and FNR. As can be seen, \geminipro was able to give the best accuracy (95.5\%) and guarantee the balance between the low value of FPR (4.5\%) and the low value of FNR (4.5\%). Other LLMs that produced lower FNR have higher FPR. In general, all LLMs have good violence detection performance in videos. However, \llamavision gave the lowest accuracy (82.12\%) and the highest FNR (17.75\%) among LLMs for the violence detection task in video inputs. Additionally, small versions of LLMs, such as \geminiflash, that can usually balance speed and efficiency, also show superior performance with \geminiflash's accuracy of (94.13\%). 

For detecting violence in the video dataset, various state-of-the-art baseline methods that have used this dataset, including CNN-LSTM~\cite{aldahoul2021convolutional}, MobileNet V2-LSTM~\cite{abdullah2023combination}, DenseNet-121-LSTM~\cite{abdullah2023combination}, ShuffleNet-LSTM~\cite{abdullah2023combination}, and EfficientNet-B0~\cite{abdullah2023combination} have been used for comparison. Table~\ref{tab:visual_violence_LLMs} shows the performance metrics of these baseline methods mentioned in their main articles. We added an X in some of the cells to represent the baseline methods that did not provide the FNR. All LLMs except \llama were able to outperform the baseline methods significantly by increasing the accuracy and reducing the FPR and FNR.  

Our finding highlights the significance of using the vision capability of LLMs for violence detection in videos.  Additionally, the use case of classifying violence videos using LLMs that we showed in this work can be extended at scale to study the potential of using LLMs to classify various multi-class video datasets using the features of feeding LLMs with multiple images or frames.

\begin{table}[!htb]
\centering
\renewcommand{\arraystretch}{1.2} %
{\footnotesize
\begin{tabularx}{\columnwidth}{|X|c|c|c|c|c|c|}
\hline
                                & \textbf{Accuracy} & \textbf{Precision} & \textbf{Recall} & \textbf{F1 score} & \textbf{FPR} & \textbf{FNR} \\ \hline
\textbf{\cellcolor{yellow!30}{CNN-LSTM (baseline)~\cite{aldahoul2021convolutional}}}                  & 73.35\%             & 72.53\%              & 76.9\%          & 74.01\%            & 30.2\%        & 23.1\%        \\ \hline
\textbf{\cellcolor{yellow!30}{MobileNetV2– LSTM (baseline)~\cite{abdullah2023combination}}}            & 82.88\%          & 82.13\%              & 83.59\%           & 82.85\%             & 17.82\%        & X         \\ \hline
\textbf{\cellcolor{yellow!30}{DenseNet-121– LSTM (baseline)~\cite{abdullah2023combination}}}            & 80\%          & 79.95\%              & 79.55\%           & 79.75\%             & 19.55\%        & X         \\ \hline
\textbf{\cellcolor{yellow!30}{ShuffleNet– LSTM (baseline)~\cite{abdullah2023combination}}}            & 66.13\%          & 64.24\%              & 71.21\%           & 67.54\%             & 38.86\%        & X         \\ \hline
\textbf{\cellcolor{yellow!30}{EfficientNet-B0–LSTM (baseline~\cite{abdullah2023combination}}}            & 86.38\%          & 86.5\%              & 86.28\%           & 86.39\%             & 13.53\%        & X         \\ \hline
\textbf{\cellcolor{blue!30}{\gpt}}                  & 93.75\%             & 89.06\%              & 99.75\%           & 94.10\%             & 12.25\%        & 0.25\%         \\ \hline
\textbf{\cellcolor{blue!30}{\gptmini}}            & 91.75\%             & 86.30\%              & 99.25\%           & 92.33\%             & 15.75\%        & 0.75\%         \\ \hline
\textbf{\cellcolor{blue!30}{\geminipro} }        & 95.5\%              & 95.5\%               & 95.5\%            & 95.5\%              & 4.5\%          & 4.5\%          \\ \hline
\textbf{\cellcolor{blue!30}{\geminiflash}}       & 94.13\%             & 91.14\%              & 97.75\%           & 94.33\%             & 9.5\%          & 2.25\%         \\ \hline
\textbf{\cellcolor{blue!30}{\llamavision}} & 82.12\%             & 82.04\%              & 82.25\%           & 82.15\%             & 18\%           & 17.75\%        \\ \hline
\end{tabularx}
}
\caption{Comparison between \colorbox{blue!30}{LLMs} and \colorbox{yellow!30}{baseline} methods for the violence video classification task.}
\label{tab:visual_violence_LLMs}
\end{table}

\subsubsection{Detection of Nudity, Pornography, and Sexual acts in Images}

In the second experiment, we aim to study the LLMs capability to detect the nudity, pornography, and sexual acts in the images such as traditional human photos~\cite{aldahoul2020transfer}, cartoon~\cite{aldahoul2021evaluation}, and sketch~\cite{momo2023evaluation}. We used three datasets: human photo dataset, cartoon dataset, and sketch dataset that have been described in the dataset overview section. We used seven LLMs that support vision by considering one image at input driven by a textual prompt. 

The prompt used for nudity/pornography detection in all types of images such as human photos, cartoons, and sketches is:

\begin{tcolorbox}[colback=orange!5!white, colframe=orange!75!black, title={Nudity/Pornography prompt}, rounded corners, boxrule=1pt, boxsep=1pt]
Classify the image content into one of these classes; choose only from this list:
[`nude',`porn',`naked',`sex',`normal'], answer using a single word.
\end{tcolorbox}

~\Cref{tab:visual_human_LLMs,tab:visual_cartoon_LLMs,tab:visual_sketch_LLMs} present the results comparing various LLMs with one another, as well as with the baseline methods. For detecting nudity/pornography in the human photo dataset, various baseline methods, including ResNet 50~\cite{aldahoul2020transfer}, YOLO-TesNet50~\cite{aldahoul2020transfer}, and cloud-based solutions like AWS~\cite{aldahoulevaluation} and Microsoft Azure~\cite{aldahoulevaluation} have been used for comparison.

\begin{table}[!htb]
\centering
\renewcommand{\arraystretch}{1.2} %
{\footnotesize
\begin{tabularx}{\columnwidth}{|X|c|c|c|c|c|c|}
\hline
                                & \textbf{Accuracy} & \textbf{Precision} & \textbf{Recall} & \textbf{F1 score} & \textbf{FPR} & \textbf{FNR} \\ \hline

\textbf{\cellcolor{yellow!30}{ResNet50 (baseline)~\cite{aldahoul2020transfer}}}                  & 85.5\%            & 85.55\%              & 85.75\%          & 85.57\%             & 14.75\%        & 14.25\%         \\ \hline

\textbf{\cellcolor{yellow!30}{YOLO-ResNet50 (baseline)~\cite{aldahoul2020transfer}}}                  & 89.5\%             & 91\%              & 87.75\%           & 89.27\%             & 8.75\%        & 12.25\%         \\ \hline
\textbf{\cellcolor{yellow!30}{AWS (baseline)~\cite{aldahoulevaluation}}}  & 84\%    & 85\%                & 84\%              & 84\%             & x            & x           \\ \hline

\textbf{\cellcolor{yellow!30}{Microsoft Azure (baseline)~\cite{aldahoulevaluation}}}                  & 87\%           & 89\%                & 87\%              & 87\%             & x            & x            \\ \hline

\textbf{\cellcolor{blue!30}{\gpt}}                  & 96\%                & 100\%                & 92\%              & 95.83\%             & 0\%            & 8\%            \\ \hline
\textbf{\cellcolor{blue!30}{\gptmini}}             & 91\%                & 100\%                & 82\%              & 90.11\%             & 0\%            & 18\%           \\ \hline
\textbf{\cellcolor{blue!30}{\geminipro}}         & 93.13\%             & 99.15\%              & 87\%              & 92.68\%             & 0.75\%         & 13\%           \\ \hline
\textbf{\cellcolor{blue!30}{\geminiflash}}       & 95.37\%             & 100\%                & 90.75\%           & 95.15\%             & 0\%            & 9.25\%         \\ \hline
\textbf{\cellcolor{blue!30} {\llamavision}} & 97.63\%             & 99.48\%              & 95.75\%           & 97.58\%             & 0.5\%          & 4.25\%         \\ \hline
\textbf{\cellcolor{blue!30}{\llamaguardvision}}    & 89.38\%             & 94.87\%              & 83.25\%           & 88.68\%             & 4.5\%          & 16.75\%        \\ \hline
\textbf{\cellcolor{blue!30}{\openaimod}}   & 83.5\%   & 100\%     & 67\%      & 80.24\%      & 0\%     & 33\%           \\ \hline
\end{tabularx}
}
\caption{Comparison between \colorbox{blue!30}{LLMs} and \colorbox{yellow!30}{baseline} methods for traditional photos of human nudity/pornography detection task.}
\label{tab:visual_human_LLMs}
\end{table}

In Table~\ref{tab:visual_human_LLMs}, both \llamaguardvision and \gpt provide significant improvements over LLMs and baseline methods, particularly in achieving the highest accuracy and maintaining a balance between precision and recall. Table~\ref{tab:visual_human_LLMs} also highlights the  superiority of \llamavision having the best accuracy of 97.63\% and capturing more true positives, making it suitable for cases where high FNR is critical. 

We also observe that \gptmini has high precision, meaning it rarely generates false positives, but it sacrifices recall, leading to more false negatives. Additionally, both \openaimod and \llamaguardvision gave the lowest accuracy and the highest FNR among LLMs for this task of nudity, pornography, and sexual act detection in human photos, which makes them not the optimal candidate models for this task, answering \textbf{RQ1}. Even \openaimod is accurate in flagging normal content, showing a precision of 100\%, its recall is lower (67\%), indicating it misses many true positives, as reflected in the higher FNR (33\%).

It is also worth mentioning that \geminiflash maintains a high accuracy of 95.37\%, with strong precision (100\%) and a balanced recall. This combination of speed and accuracy makes it effective for situations where both quick responses and reliable content filtering are required, such as live-streaming platforms. 

Our finding underscores the crucial role of leveraging the vision capabilities of LLMs in detecting sensitive content, such as nudity, pornography, and sexual acts, in both static images and dynamic videoframes, which is essential for large-scale and real-time applications adhering to safety guidelines.

\begin{table}[!htb]
\centering
\renewcommand{\arraystretch}{1.2} %
{\footnotesize
\begin{tabularx}{\columnwidth}{|X|c|c|c|c|c|c|}
\hline
                       & \textbf{Accuracy} & \textbf{Precision} & \textbf{Recall} & \textbf{F1 score} & \textbf{FPR} & \textbf{FNR}  \\ \hline
\textbf{\cellcolor{yellow!30}{ResNet50 (baseline)~\cite{aldahoul2021evaluation}}} &94.11\%    &93.7\%	&93.25\%  &93.42\% 	&5.2\%	&6.75\%		           \\ \hline
\textbf{\cellcolor{yellow!30}{ResNet101 (baseline)~\cite{aldahoul2021evaluation}}}     &94.22\% &93.85\%	&93.25\%		&93.52\% &5\%  &6.75\%		           \\ \hline
\textbf{\cellcolor{yellow!30}{EfficientNet-B0 (baseline)~\cite{aldahoul2021evaluation}}}   &94\%   &94.13\%	&92.25\%		&93.18\%     &4.60\%   &7.75\%		            \\ \hline
\textbf{\cellcolor{yellow!30}{EfficientNet-B4 (baseline)~\cite{aldahoul2021evaluation}}}   &92.78\%	&93.4\%   &90.25\%	&91.77\%   &5.2\%  &9.75\%	           \\ \hline
\textbf{\cellcolor{yellow!30}{EfficientNet-B7 (baseline)~\cite{aldahoul2021evaluation}}}   &94.22\%	&93.69\%	   &93.5\%	&93.54\%   &5.2\%   &6.5\%		           \\ \hline

\textbf{\cellcolor{yellow!30}{AWS (baseline)~\cite{aldahoulevaluation}}}                  & 97\%                & 97\%               & 97\%              & 97\%            & x            & x            \\ \hline
\textbf{\cellcolor{yellow!30}{Microsoft Azure (baseline)~\cite{aldahoulevaluation}}}                  & 97\%                & 97\%                & 97\%              & 97\%             & x            & x            \\ \hline
\textbf{\cellcolor{blue!30}{\gpt}}                  & 98.67\%    & 100\%       & 97\%     & 98.48\%    & 0\%   & 3\%    \\ \hline
\textbf{\cellcolor{blue!30}{\gptmini}}             & 97\%       & 100\%       & 93.25\%  & 96.51\%    & 0\%   & 6.75\% \\ \hline
\textbf{\cellcolor{blue!30}{\geminipro}}         & 99.89\%    & 99.75\%     & 100\%    & 99.87\%    & 0.2\% & 0\%    \\ \hline
\textbf{\cellcolor{blue!30}{\geminiflash}}       & 99.44\%    & 100\%       & 98.75\%  & 99.37\%    & 0\%   & 1.25\% \\ \hline
\textbf{\cellcolor{blue!30}{\llamavision}} & 98.78\%    & 98.50\%     & 98.75\%  & 98.63\%    & 1.2\% & 1.25\% \\ \hline
\textbf{\cellcolor{blue!30}{\llamaguardvision}}     & 98.56\%    & 99.49\%     & 97.25\%  & 98.36\%    & 0.4\% & 2.75\% \\ \hline
\textbf{\cellcolor{blue!30}{\openaimod}}      & 96.88\%    & 100\%       & 93\%     & 96.37\%    & 0\%   & 7\%    \\ \hline
\end{tabularx}
}
\caption{Comparison between \colorbox{blue!30}{LLMs} and \colorbox{yellow!30}{baseline} methods for cartoon image nudity/pornography detection task.}
\label{tab:visual_cartoon_LLMs}
\end{table}

For detecting nudity/pornography in cartoon image dataset, various baseline methods, including various versions of EfficientNet (B0, B4, B7)~\cite{aldahoul2021evaluation}, ResNet (50 and 101)~\cite{aldahoul2021evaluation}, and cloud-based solutions like AWS~\cite{aldahoulevaluation} and Microsoft Azure~\cite{aldahoulevaluation} have been used for comparison. 

Table~\ref{tab:visual_cartoon_LLMs} suggests that LLMs, with their advanced vision capabilities, outperform traditional CNN-based baselines in terms of precision and recall, making them suitable for applications where accurate detection is crucial. Table~\ref{tab:visual_cartoon_LLMs} highlights the dominance of \geminiflash and \geminipro with their near-perfect performance across all metrics, highlighting their efficiency in both FPR and FNR, indicating fewer misclassifications, which is crucial for tasks that require high reliability.

\begin{table}[!htb]
\centering
\renewcommand{\arraystretch}{1.2} %
{\footnotesize
\begin{tabularx}{\columnwidth}{|X|c|c|c|c|c|c|}
\hline
\textbf{}                       & \textbf{Accuracy} & \textbf{Precision} & \textbf{Recall} & \textbf{F1 score} & \textbf{FPR} & \textbf{FNR}  \\ \hline

\textbf{\cellcolor{yellow!30}{CoAtNet0 (baseline)~\cite{momo2023evaluation}}}    &94\%        & 94\%          & 94\%              & 94\%     & x            & x               \\ \hline
\textbf{\cellcolor{yellow!30}{AWS (baseline)~\cite{aldahoulevaluation}}}         &93\%        & 93\%          & 93\%              & 93\%       & x  & x                   \\ \hline
\textbf{\cellcolor{yellow!30}{Microsoft Azure (baseline)~\cite{aldahoulevaluation}}}         &81\%        & 84\%          & 81\%              & 81\%       & x  & x                   \\ \hline
\textbf{\cellcolor{blue!30}{\gpt}}                  & 95.4\%     & 98.92\%     & 91.8\%   & 95.23\%    & 1\%   & 8.2\%  \\ \hline
\textbf{\cellcolor{blue!30}{\gptmini}}             & 93.7\%     & 99.55\%     & 87.8\%   & 93.31\%    & 0.4\% & 12.2\%\% \\ \hline
\textbf{\cellcolor{blue!30}{\geminipro}}         & 94.6\%     & 99.56\%     & 89.6\%   & 94.31\%    & 0.4\% & 10.4\% \\ \hline
\textbf{\cellcolor{blue!30}{\geminiflash}}       & 95.2\%     & 99.13\%     & 91.20\%  & 95\%       & 0.8\% & 8.8\%  \\ \hline
\textbf{\cellcolor{blue!30}{\llamavision}} & 87.8\%     & 93.95\%     & 80.8\%   & 86.88\%    & 5.2\% & 19.2\% \\ \hline
\textbf{\cellcolor{blue!30}{\llamaguardvision}}     & 88.8\%     & 97.78\%     & 79.4\%   & 87.64\%    & 1.8\% & 20.6\% \\ \hline
\textbf{\cellcolor{blue!30}{\openaimod}}      & 83.7\%     & 100\%       & 67.4\%   & 80.53\%    & 0\%   & 32.6\% \\ \hline
\end{tabularx}
}
\caption{Comparison between \colorbox{blue!30}{LLMs} and \colorbox{yellow!30}{baseline} methods for sketch image nudity/pornography detection task.}
\label{tab:visual_sketch_LLMs}
\end{table}

Both Llama guard and openAI moderation models have already been trained for content moderation tasks to detect violence, hate, harm, and sexual contents. The superior performance of Llama 3 Guard and the openAI moderation model in detecting nudity or pornography in cartoon images compared to human photos may indicate that these LLMs have a strong ability to identify explicit content within cartoon datasets but may struggle to detect general nudity in datasets containing human photos.

For detecting nudity/pornography in the sketch image dataset, various baseline methods, including CoAtNet0~\cite{momo2023evaluation}, and cloud-based solutions like AWS~\cite{aldahoulevaluation} and Microsoft Azure~\cite{aldahoulevaluation} have been used for comparison. 

Table~\ref{tab:visual_sketch_LLMs} shows
the highest accuracy in \gpt (95.4\%), and \geminiflash (95.2\%) among all models. It also highlights low FNR in both LLMs, which makes them perfect for applications needing high sensitivity.

openAI moderation model's has perfect precision and zero FPR, which indicates it's very conservative, preferring to classify fewer cases as positive to avoid false positives, but at the cost of missing many true positives with very high FNR (32.6\%). Additionally, both Llama 3 instruct and Llama 3 guard lag behind in recall and suffer from higher FNR. In summary, both \openaimod and \llamaguardtext show limitations in detecting pornography and sex acts in sketch images, which addresses \textbf{RQ1}.

\subsubsection{Detection of Graphic Violence in Images}

The third experiment seeks to assess the LLMs capability to detect graphic violence in images. We used the graphic violence dataset that has been described in the dataset overview section. We explored seven LLMs that support vision by considering one image at input driven by a textual prompt. The prompt used for bloody content detection in the images is:

\begin{tcolorbox}[colback=orange!5!white, colframe=orange!75!black, title={Graphic Violence prompt}, rounded corners, boxrule=1pt, boxsep=1pt]
Classify the image content into one of these classes; choose only from this list: [`violence',`graphic violence',`normal'], answer using a single word.
\end{tcolorbox}

The results in Table~\ref{tab:visual_abuse_LLMs} underscore that \llamaguard is not able to graphic violence content in the images, which should be considered under the `graphic violence' category. The \llamaguard has two categories related to `suicide and self-harm' and `violent crimes', but it only generates the category of `intellectual property', which refers to prompts or responses that may violate the intellectual property rights of any third party, for the majority of images in the bloody content dataset. 

\begin{table}[!htb]
\centering
\renewcommand{\arraystretch}{1.2} %
{\footnotesize
\begin{tabularx}{\columnwidth}{|X|c|c|c|c|}
\hline
\textbf{}                       & \textbf{Graphic} & \textbf{Alcohol} & \textbf{Drug} & \textbf{Child} \\
\textbf{}                       & \textbf{violence} & \textbf{abuse} & \textbf{abuse} & \textbf{abuse} \\ \hline

\textbf{\cellcolor{blue!30}{\gpt}}     & 92.86\% & 94.77\%     &91.67\% & 81.75\%
    \\ \hline
\textbf{\cellcolor{blue!30}{\gptmini}}  & 96.57\%   & 88.42\%      & 93.75\%    & 77\%    \\ \hline
\textbf{\cellcolor{blue!30}{\geminipro}}  & 92.29\%   & 75.16\%      & 89.58\%     & 70.63\%  \\ \hline
\textbf{\cellcolor{blue!30}{\geminiflash}}   & 96\% & 92.81\%       & 90.97\%    &73.02\%       \\ \hline
\textbf{\cellcolor{blue!30}{\llamavision}} &98.57\%      &87.58\%
&90.28\%    &45.24\%     \\ \hline
\textbf{\cellcolor{blue!30}{\llamaguardvision}}  &0\%      &0\%      &0\%    &0\%    \\ \hline
\textbf{\cellcolor{blue!30}{\openaimod}}   &75.43\%   &3.9\%        &29.17\%    &16.67\%    \\ \hline
\end{tabularx}
}
\caption{Accuracy of \colorbox{blue!30}{LLMs} for graphic violence, alcohol, drug, and child abuse detection task.}
\label{tab:visual_abuse_LLMs}
\end{table}

The results show that \llamavision excels in detecting graphic violence in this dataset with the highest accuracy of 98.57\%. Additionally, both \geminiflash and \gptmini show remarkable performance with an accuracy around 96\%. On the other hand, \openaimod shows a moderate detection rate of 75.43\%, which is lower than all other models listed except \llamaguard. This suggests that \openaimod might have limitations in recognizing graphic violence considering \openaimod has been fine-tuned to generate a predefined list of categories including `violence' (which depicts death, violence, or physical injury), and `self-harm' (which encourages or depicts acts of self-harm, such as suicide and cutting). Our finding highlights a gap in performance of \openaimod and \llamaguard in this specific task of graphic violence detection which addresses \textbf{RQ1} and we show that other LLMs such as \gpt, \gemini, \llama are able to overcome these limitations which addresses \textbf{RQ2}.

For the aforementioned tasks of detecting violence in videos, nudity, and pornography in human photos, cartoons, and sketches, and graphic violence in images, we can see that LLMs exhibit inconsistency in prediction accuracy across different types of inappropriate content, such as  violence and adult content, thus answering \textbf{RQ3}.

\subsubsection{Detection of Alcohol, Drug, and Child Abuse in Images}

The fourth experiment intends to evaluate the LLMs capability to detect alcohol, drug, and child abuse in the images. We utilized a dataset of images that have various abuse contents described in the dataset overview section. We examined seven vision-enabled LLMs, each capable of processing a single image input guided by a textual prompt. The prompt used for abuse detection in the images is:

\begin{tcolorbox}[colback=orange!5!white, colframe=orange!75!black, title={Abuse prompt}, rounded corners, boxrule=1pt, boxsep=1pt]
Classify the image content into one of these classes; choose only from this list:
[`child abuse',`alcohol abuse',`drug abuse',`normal'], answer using a single word.
\end{tcolorbox}

It is clear from Table~\ref{tab:visual_abuse_LLMs} that \gpt is the best at achieving the highest accuracy in the categories of `alcohol abuse' and `child abuse'. Similarly, \gptmini gave the top accuracy for drug abuse detection. Additionally, \geminiflash performs well in detecting `drug abuse' and `alcohol abuse' categories with scores of 90.97\% and 92.81\%, respectively. However, most models, and particularly \llamavision, struggle with detecting `child abuse' category, indicating a need for improved sensitivity. On the other hand, the detection capability of \openaimod for content related to alcohol abuse, drug abuse, and child abuse is minimal because the abuse category is not related to a pre-defined list of categories that \openaimod has been fine-tuned to generate. Similarly, \llamaguard has no recorded accuracy for any abuse category. This indicates that both aforementioned \openaimod and \llamaguard lack the capability to detect these forms of inappropriate content, which addresses \textbf{RQ1}. Additionally, we show that other LLMs, such as \gpt, \gemini, and \llama are able to overcome these limitations, which address \textbf{RQ2}.

\subsection{Textual Content Analysis}

We carried out several experiments to demonstrate the potential of using LLMs, such as \llamaguard, \openaimod, \gpt,\gemini, and \llama for textual content censorship and particularly violence, sexual text, and hate speech. Several datasets, such as violence tweets, hate speech tweets, and adult content articles' descriptions, have been utilized in these experiments to evaluate various LLMs. 

Additionally, we exploited these LLMs with two use cases: The first use case is articles from five news outlets aiming to identify whether the language used contains implicit or explicit hate speech, violence, or adult content, helping to filter content before publication or alert readers to sensitive topics. The second use case is Amazon beauty reviews, aiming to exploit LLMs in assessing customers' feedback to ensure that reviews remain respectful and adhere to guidelines, enhancing the platform's safety and trustworthiness.

As \openaimod was already trained for moderation tasks, the API is designed without the need to feed it by any prompt. We feed the text required censoring to \openaimod and it generates a list of pre-defined categories associated with confidence scores. Similarly, \llamaguard is already trained for content moderation to check if there is inappropriate content in prompt, response, or chat  and generate one of a pre-defined list of categories. The \llamaguard was fed by text required censoring, and the prompt used is as follows: `Check if there is unsafe content in the following text'. The prompt used for \gpt and \gemini is:

\begin{tcolorbox}[colback=orange!5!white, colframe=orange!75!black, title={\gpt and \gemini prompt}, rounded corners, boxrule=1pt, boxsep=1pt]
Find the category and its confidence score (between 0 and 10) of this text based on the following list:
[`harassment',`hate',`harm',`sexual',`violence',`graphic violence',`normal']; answer using this format \{`category':category,`score':score\}.
\end{tcolorbox}

The prompt used with \llama is as follows:

\begin{tcolorbox}[colback=orange!5!white, colframe=orange!75!black, title={\llama prompt}, rounded corners, boxrule=1pt, boxsep=1pt]
Task: check if there is unsafe content in this text containing requests according to our safety policy with the below categories.
\\<BEGIN UNSAFE CONTENT CATEGORIES> 
\\`normal'
\\`violence'
\\`graphic violence'
\\`harassment'
\\`hate'
\\`harm'
\\`sexual'
\\<END UNSAFE CONTENT CATEGORIES>
\\Provide your safety assessment for this text: The first line must read `safe' or `unsafe'.
If unsafe, a second line must include the assessed safety category from the above categories only and a confidence score separated by a comma. 
\end{tcolorbox}

For \gemini, we set the safety settings to be `BLOCK-LOW-AND-ABOVE'. In this case, \gemini will block prompts when the probability score is either low, medium, or high. In this experiment, the text samples that were blocked with medium and high probability scores are considered to have adult content. On the other hand, the samples that were blocked with low probability scores and the unblocked samples are predicted under the `normal' category. For \gpt, due to the lack of safety settings (unlike \gemini), the text samples were predicted to be under one of the categories defined in the \gpt and \gemini prompts. 

\subsubsection{Detection of Adult Content in Articles' Description}

In this experiment, we used the adult content dataset~\cite{Adult-content-dataset} mentioned in the data overview section to evaluate the capabilities of LLMs for adult content censorship. Seven LLMs have been examined to process the text of the articles' descriptions that may contain adult content. The LLMs are driven by the above prompts for detection.

For \llama, in the adult content detection of the articles' descriptions, the `harassment' category was generated for several samples; even the `sexual' category is available as an option in the \llama's prompt. If this `harassment' category is considered an 'adult' category, the performance metrics of \llama are shown in Table~\ref{tab:texual_adult_LLMs}. If we consider the `harassment' category as a 'non-adult' category because the model fails to select the `sexual' category from the options given, the accuracy would degrade largely to 81.41\% with a low recall of 53.55\% and a high FNR of 46.44\%. 

Both \gpt and \geminipro achieve the highest overall accuracy and F1 scores (97.18\% and 96.34\% for \gpt, 97.06\% and 96.22\% for \geminipro). Additionally, both are optimal solutions for this task, providing balanced precision-recall and balanced FPR-FNR.

\gptmini has a noticeable drop in recall (80.47\%) compared to \gpt, resulting in a lower F1 score and higher FNR (19.5\%), which might make it less suitable for cases where FNR is critical.

\openaimod excels in precision (100\%) but has a significantly lower recall (79.88\%) and higher FNR (20.12\%). Similarly, \llamaguardtext has high precision (99.66\%) but a moderate recall (87.87\%), suggesting a similar trend to \openaimod, albeit with better recall and FNR.

\llamatext has the highest FPR (6.05\%), which could lead to more frequent false positives, potentially impacting usability depending on the context.

In summary, \llamatext and \geminipro are better suited for applications where missing a case is more problematic than an occasional false alarm. For example, in safety-critical systems, catching every instance is vital, even if it results in a few extra false positives.

\begin{table}[!htb]
\centering
\renewcommand{\arraystretch}{1.2} %
{\footnotesize
\begin{tabularx}{\columnwidth}{|X|c|c|c|c|c|c|}
\hline
\textbf{}                       & \textbf{Accuracy} & \textbf{Precision} & \textbf{Recall} & \textbf{F1 score} & \textbf{FPR} & \textbf{FNR}  \\ \hline
\textbf{\cellcolor{blue!30}{\gpt}}                  & 97.18\%     &99.37 \%     & 93.49\%   & 96.34\%    & 0.39\% & 6.51\% \\ \hline 
\textbf{\cellcolor{blue!30}{\gptmini}}             &92.12\%      &99.63\%     &80.47\%    &89.03\%     &0.20\%  &19.5\%  \\ \hline
\textbf{\cellcolor{blue!30}{\geminipro}}      & 97.06\%     & 98.45\%     & 94.08\%   & 96.22\%    & 0.98\% & 5.92\% \\ \hline
\textbf{\cellcolor{blue!30}{\geminiflash}}        & 96\%     & 98.41\%     & 91.42\%   & 94.79\%    & 0.98\% & 8.58\% \\ \hline
\textbf{\cellcolor{blue!30}{\llamatext}}  
  &94.24\%      &91.17\%      &94.67\%    &92.89\%     &6.05\%  &5.33\%  \\ \hline
\textbf{\cellcolor{blue!30}{\llamaguardtext}}   & 95.06\%     &99.66\%     & 87.87\%   & 93.40\%    &0.20\% & 12.13\% \\ \hline
\textbf{\cellcolor{blue!30}{\openaimod}}       &92.0\%       &100\%     &79.88\%    &88.82\%     &0\%  &20.12\%  \\ \hline
\end{tabularx}
}
\caption{Comparison between \colorbox{blue!30}{LLMs} for adult content in the articles' description task.}
\label{tab:texual_adult_LLMs}
\end{table}

\subsubsection{Detection of Hate Speech and Offensive Language in Tweets}

\begin{figure}[!htb]
    \centering
    \includegraphics[width=1\linewidth]{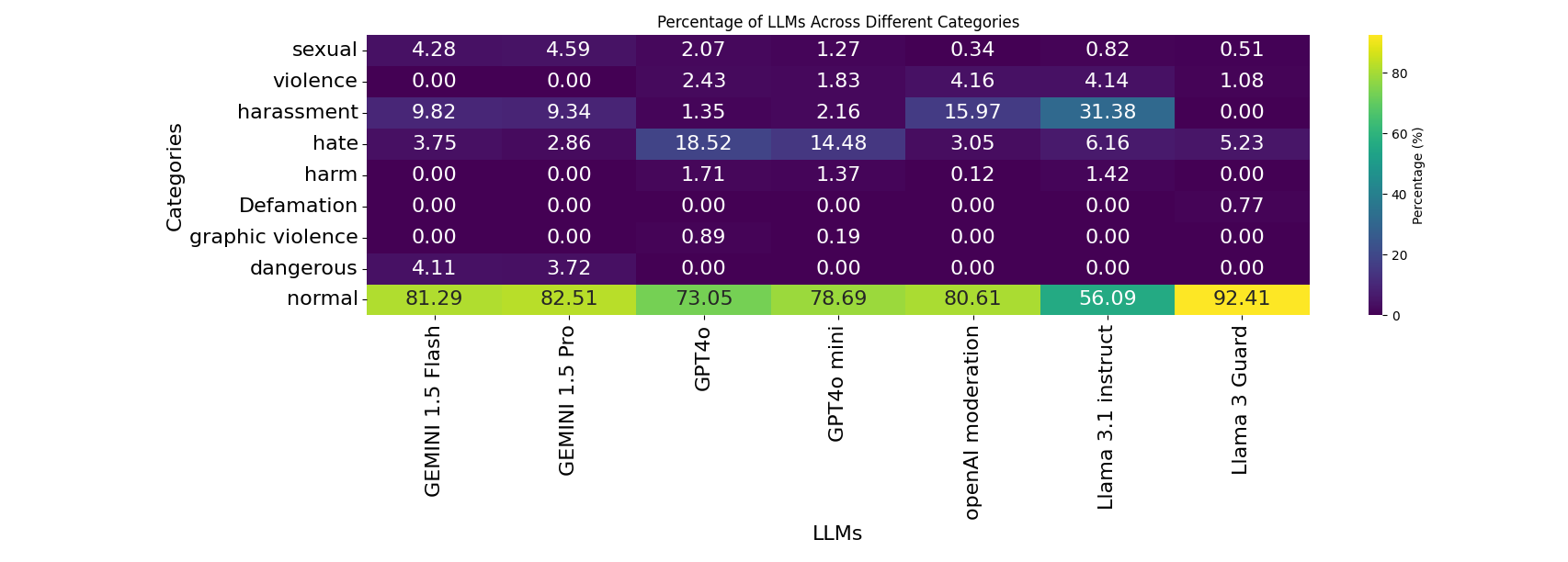}
    \caption{Heatmap of normal text showing predictions of various LLMs.}
    \label{fig:heatmap_normal}
\end{figure}

\begin{figure}[!htb]
    \centering
    \includegraphics[width=1\linewidth]{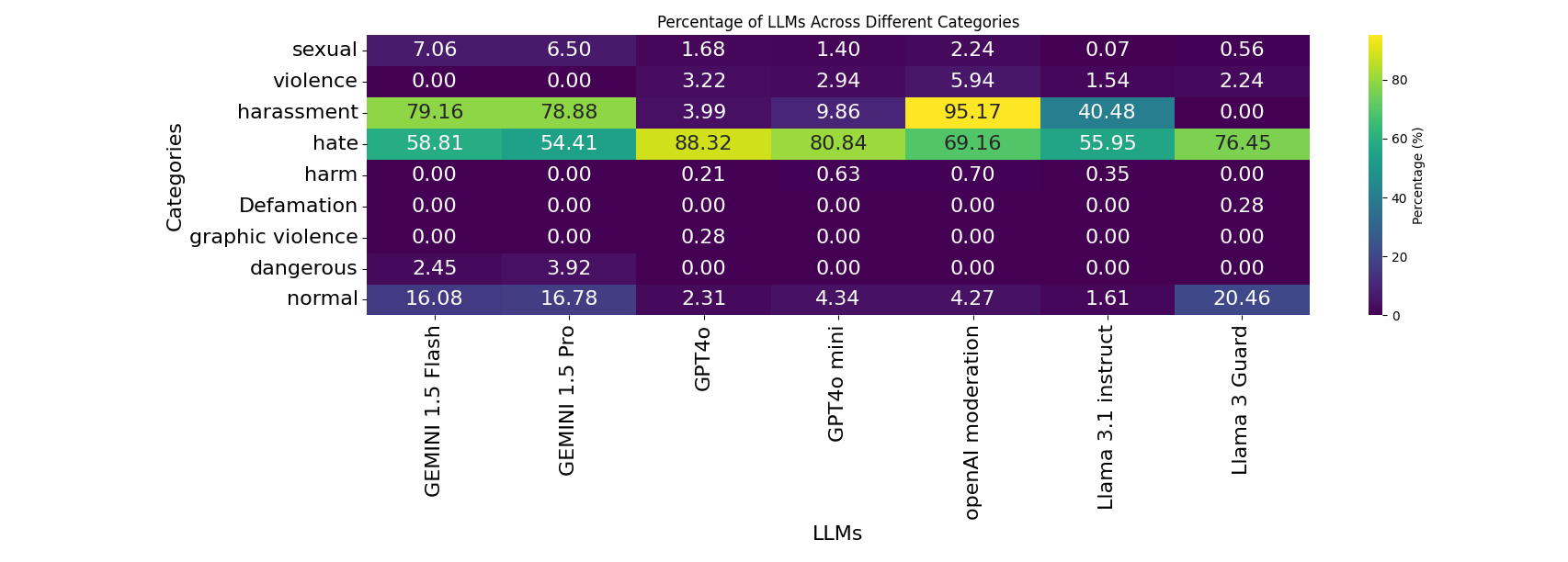}
    \caption{Heatmap of hate speech showing predictions of various LLMs.}
    \label{fig:heatmap_speech}
\end{figure}

\begin{figure}[!htb]
    \centering
    \includegraphics[width=1\linewidth]{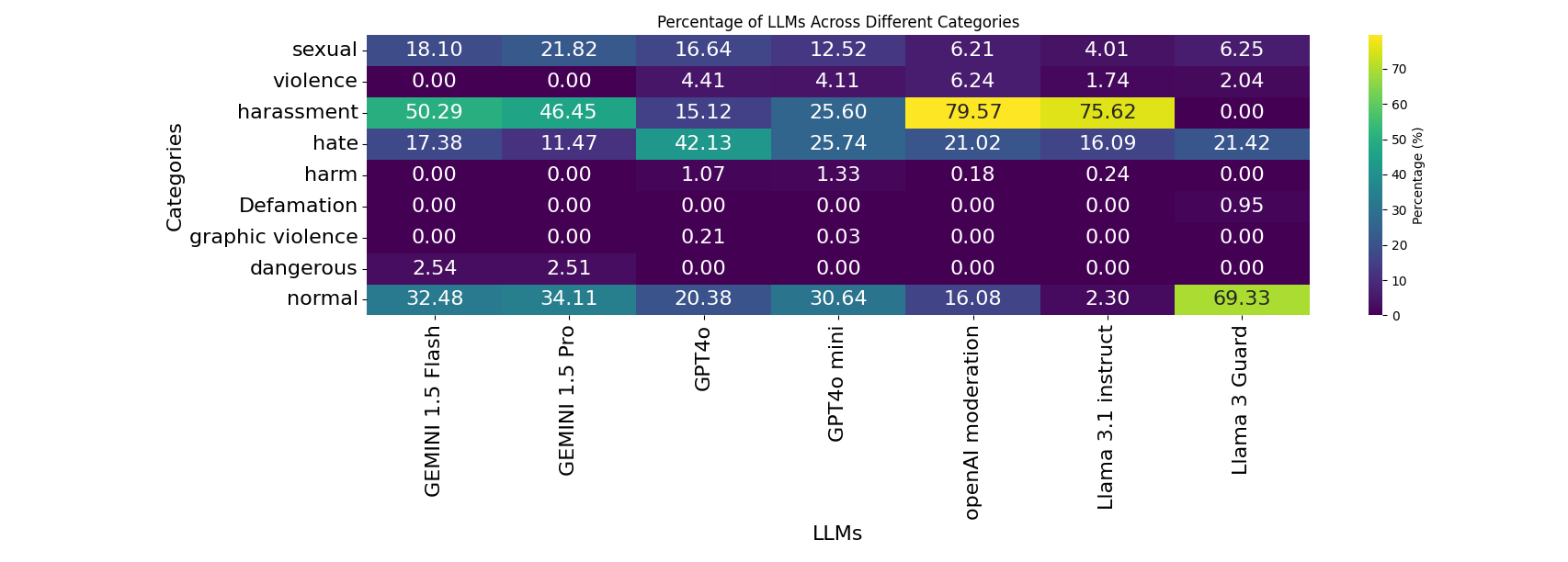}
    \caption{Heatmap of offensive language showing predictions of various LLMs.}
    \label{fig:heatmap_offensive}
\end{figure}

In this experiment, we used the hate speech and offensive language dataset~\cite{davidson2017automated,hate_speech_offensive} mentioned in the data overview section to evaluate the capabilities of LLMs for hate speech censorship. Seven LLMs have been tested with X tweets that contain hate speech and offensive language. The LLMs are driven by the prompts mentioned above for the detection. ~\Cref{fig:heatmap_normal,fig:heatmap_speech,fig:heatmap_offensive} show the results of the comparison.

We analyzed the performance of various language models in predicting hate speech, offensive language, and normal content. Across hate speech predictions, the models tend to achieve relatively high scores in categories like ``harassment'' and ``hate.'' However, the models that excel in identifying harmful content categories like hate and harassment often show reduced accuracy in identifying normal content that is neither hate speech nor offensive language, which may indicate an over-tendency to flag content as offensive.

When the high accuracy in detecting hate speech is crucial, \openaimod, \gpt, and \gptmini are optimal. However, \gpt outperforms \gptmini with lower false negative errors in offensive language classes.

\openaimod, \geminiflash, and \geminipro are able to generate multiple categories for the same text sample. In this experiment, they generate both `hate' and `harassment' in several samples labeled by the `hate speech' category. On the other hand, when \gpt, and \gemini were guided by the prompt to generate only one category, and they generated a `hate' category in this task.

Even in the case of \llamaguardtext, though it is generally so accurate in normal content detection, it has higher false negative errors in detecting hate speech and offensive language by predicting the `normal' category with 20.46\% and 69.33\%. This indicates the inability of Llama guard to detect offensive language categories in this dataset. This may be because \llamaguardtext has the `hate' category but does not have the general `harassment' category. 

Even \llamatext shows superior performance in detecting `hate speech' and `offensive language' categories, it suffers from high false positive error because it detects the `normal' category as `harassment' with 31.38\%.

\openaimod, \geminiflash, and \geminipro show the highest accuracy (above 80\%) in detecting normal content. However, both \geminiflash, and \geminipro have high false positive errors in the hate speech category (16.08\%, 16.78\%) and the offensive language category (32.48\%, 34.11\% ). On the other hand, \openaimod has lower false negative errors in both hate speech and offensive language categories. 

Overall, \openaimod shows the best capability for the task of hate and offensive language detection because it can balance false negative and false positive rates. However, it detects several instances of `hate speech' as falling under the `harassment' category.

\subsubsection{Detection of Violence in Tweets}

In this experiment, we used the violence tweet dataset~\cite{Gender-Based_Violence} mentioned in the data overview section to evaluate the capabilities of LLMs for violence censorship in textual data. Seven LLMs have been demonstrated to detect violent contents in the text of tweets that contain various types of violence. The LLMs are driven by the prompts mentioned above.

This experiment highlights the limitation of \llamaguard in detecting these five categories of violence by showing the worst performance and predicting the majority of tweets as `normal' or `defamation'. As such, \llamaguard will be excluded from the analysis below.

\begin{figure}[!htb]
    \centering
    \includegraphics[width=1\linewidth]{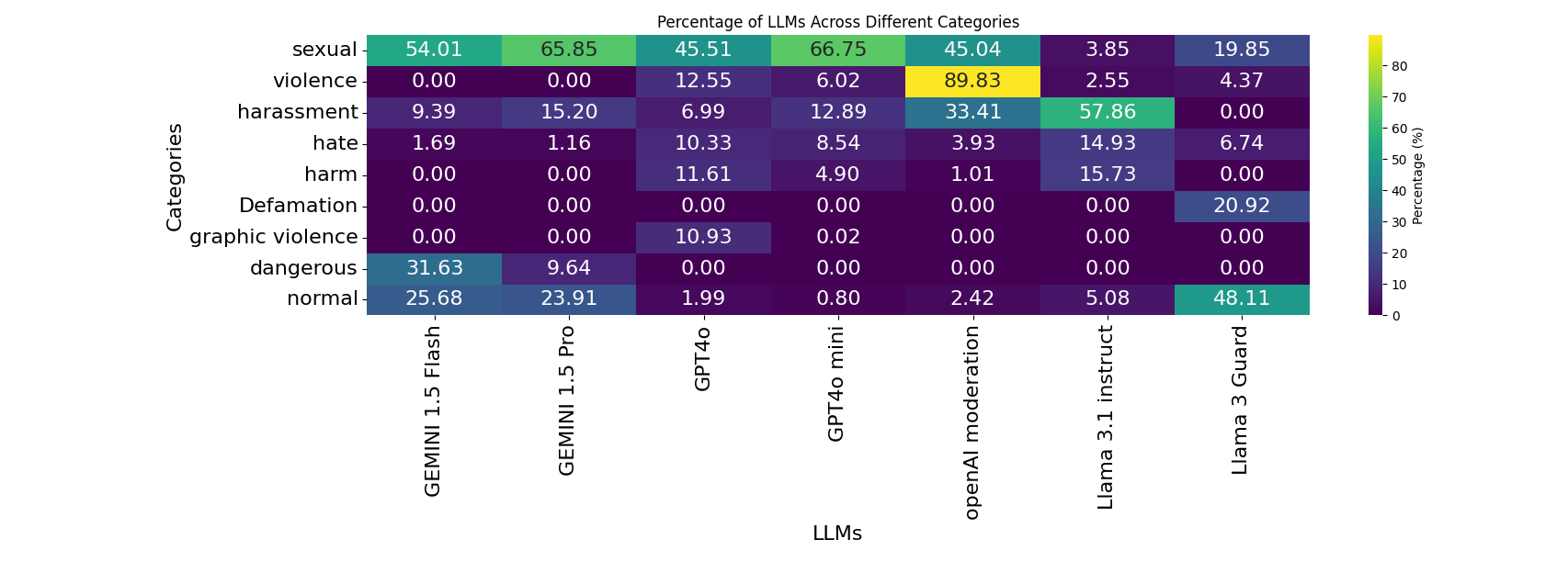}
    \caption{Heatmap of sexual violence showing predictions of various LLMs.}
    \label{fig:sexual_violence}
\end{figure}

First, we analyzed the performance of various LLMs in detecting sexual violence in labeled tweets, as shown in Figure~\ref{fig:sexual_violence}. Across the sexual violence predictions, the models tend to achieve relatively high scores in categories like `sexual' in the majority of the LLMs, `violence' in all \gpt, \gptmini, and \openaimod, `graphic violence' in \gpt, `harm' in \gpt and \llamatext and `dangerous' in \geminipro and \geminiflash. Doing an in-depth analysis, \openaimod shows high accuracy in predicting the `violence' category with 89\%. The three LLMs (\gpt, \gptmini, \openaimod) produced lower false negative error (samples predicted under the `normal' category are rare) compared to \geminipro and \geminiflash which showed a high percentage of false negative errors (23.91\% and 25.68\%). 

\begin{figure}[!htb]
    \centering
    \includegraphics[width=1\linewidth]{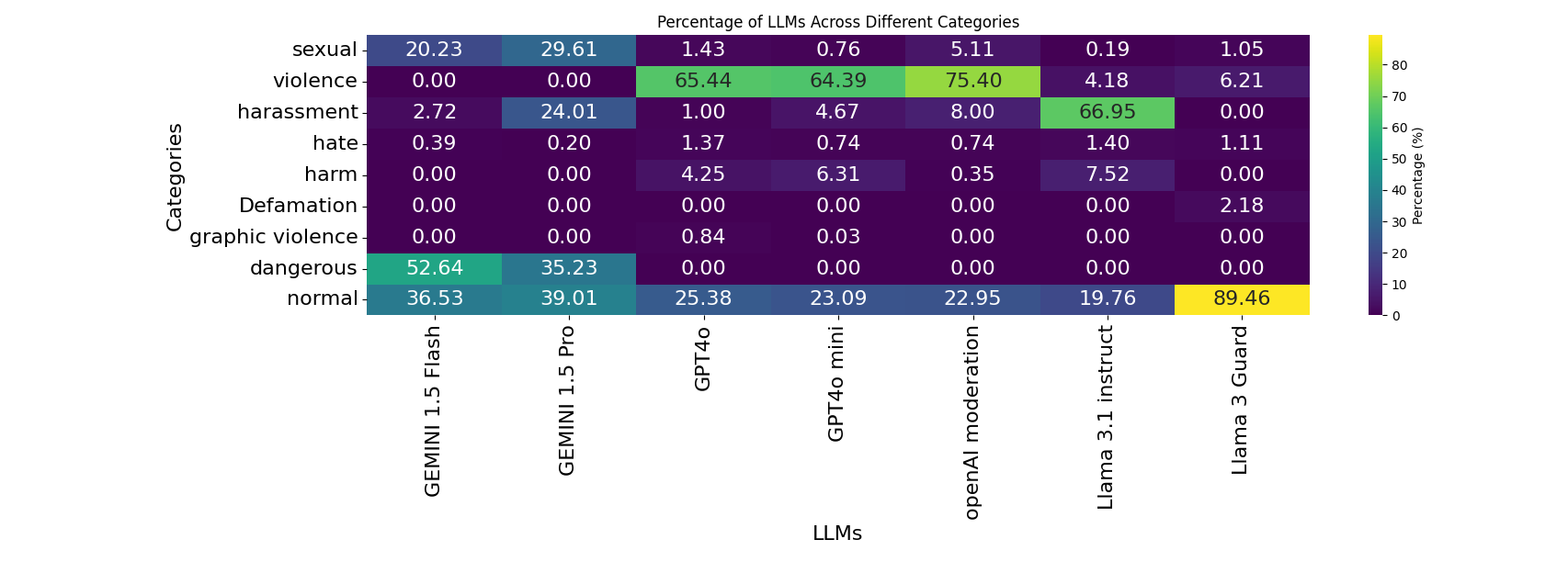}
    \caption{Heatmap of physical violence showing predictions of various LLMs.}
    \label{fig:physical_violence}
\end{figure}

Second, we evaluated the performance of various LLMs in detecting the `physical violence' in tweets, as shown in Figure~\ref{fig:physical_violence}. The difference in representing the `physical violence' between \geminipro and \geminiflash is shown as \geminipro generating more `harassment' category and less `dangerous' category than \geminiflash. Both Gemini versions produced the highest false negative errors, while other LLMs also exhibited notably high false negative errors.

\begin{figure}[!htb]
    \centering
    \includegraphics[width=1\linewidth]{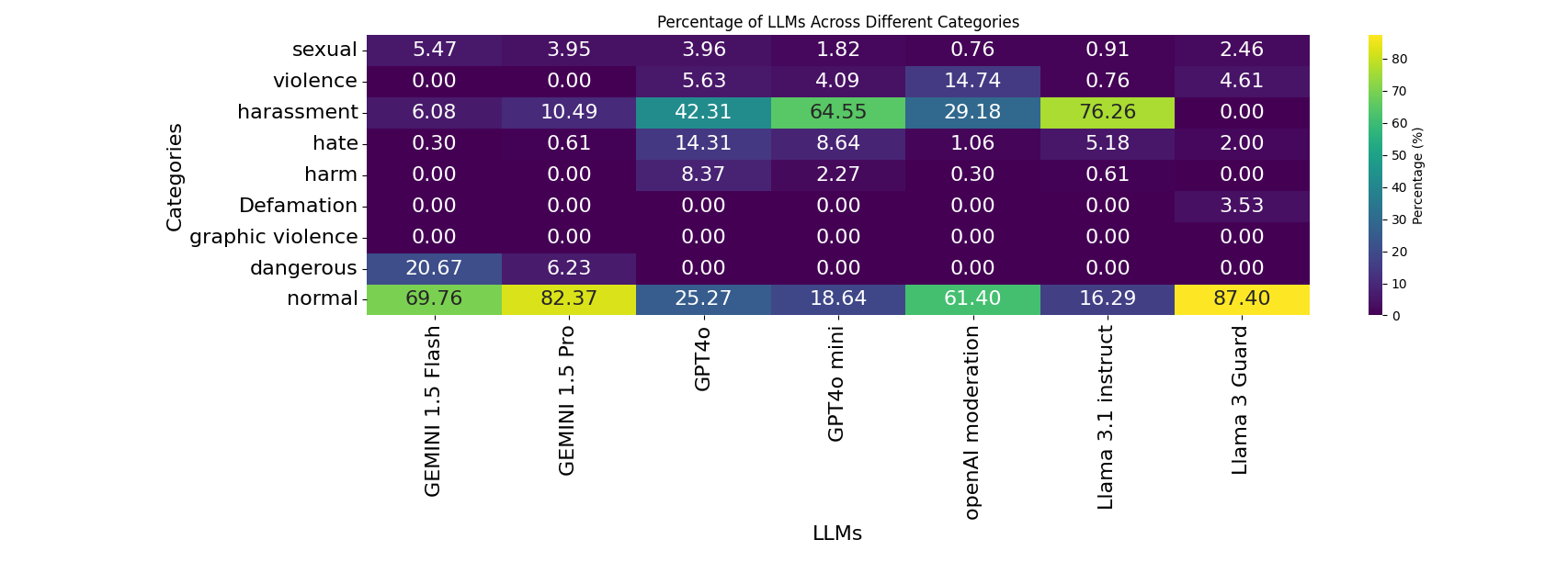}
    \caption{Heatmap of emotional violence showing predictions of various LLMs.}
    \label{fig:emotional_violence}
\end{figure}

\begin{figure}[!htb]
    \centering
    \includegraphics[width=1\linewidth]{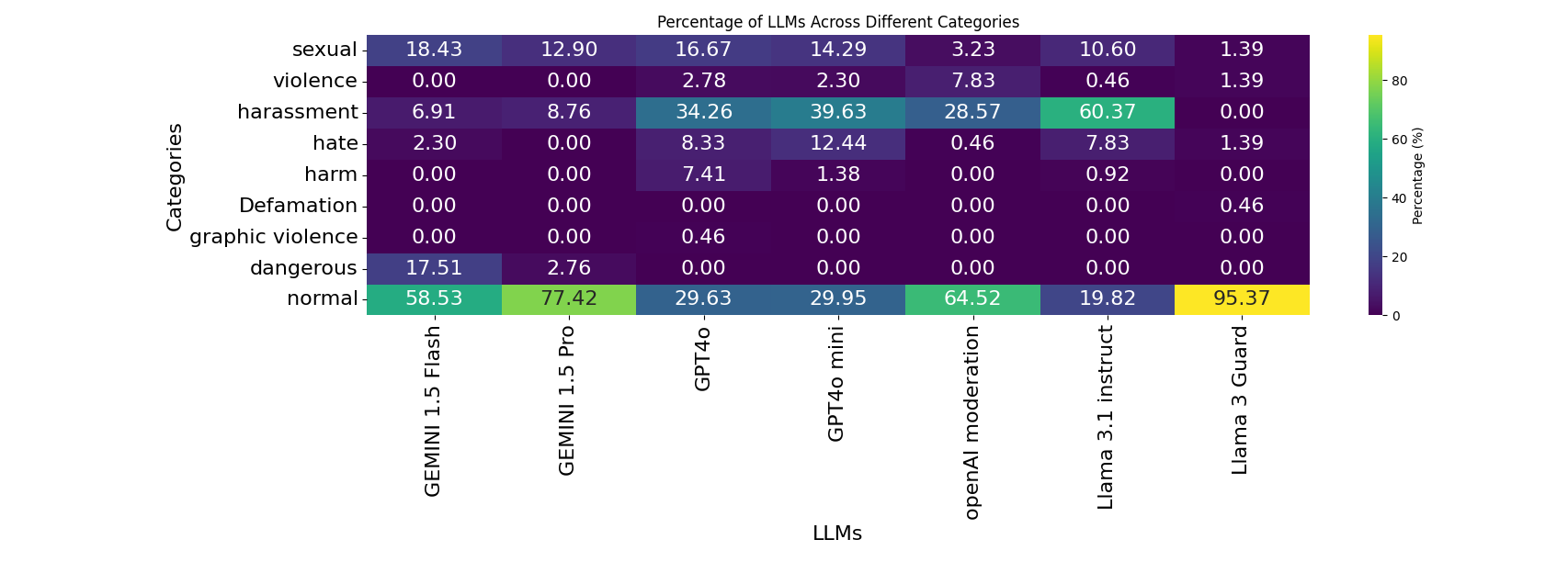}
    \caption{Heatmap of economic violence showing predictions of various LLMs.}
    \label{fig:economic_violence}
\end{figure}

Third, we assessed the performance of various LLMs in detecting the `emotional violence' and 'economic violence' in tweets, as shown in Figure~\ref{fig:emotional_violence} and Figure~\ref{fig:economic_violence}. \geminipro recorded the highest rate of false negative errors in detecting both the `emotional violence' and the `economic violence.' Additionally, both \geminiflash and \openaimod exhibited notably high false negative errors. Consequently, \gpt and \gptmini emerged as the top performers in detecting the `emotional violence' and the `economic violence', with \gptmini leading in the `emotional violence' detection by achieving the lowest false negative error rate.
This experiment indicates the limitations of \geminipro, \geminiflash, and \openaimod in detecting the `emotional violence' and the `economic violence' categories.

\begin{figure}[!htb]
    \centering
    \includegraphics[width=1\linewidth]{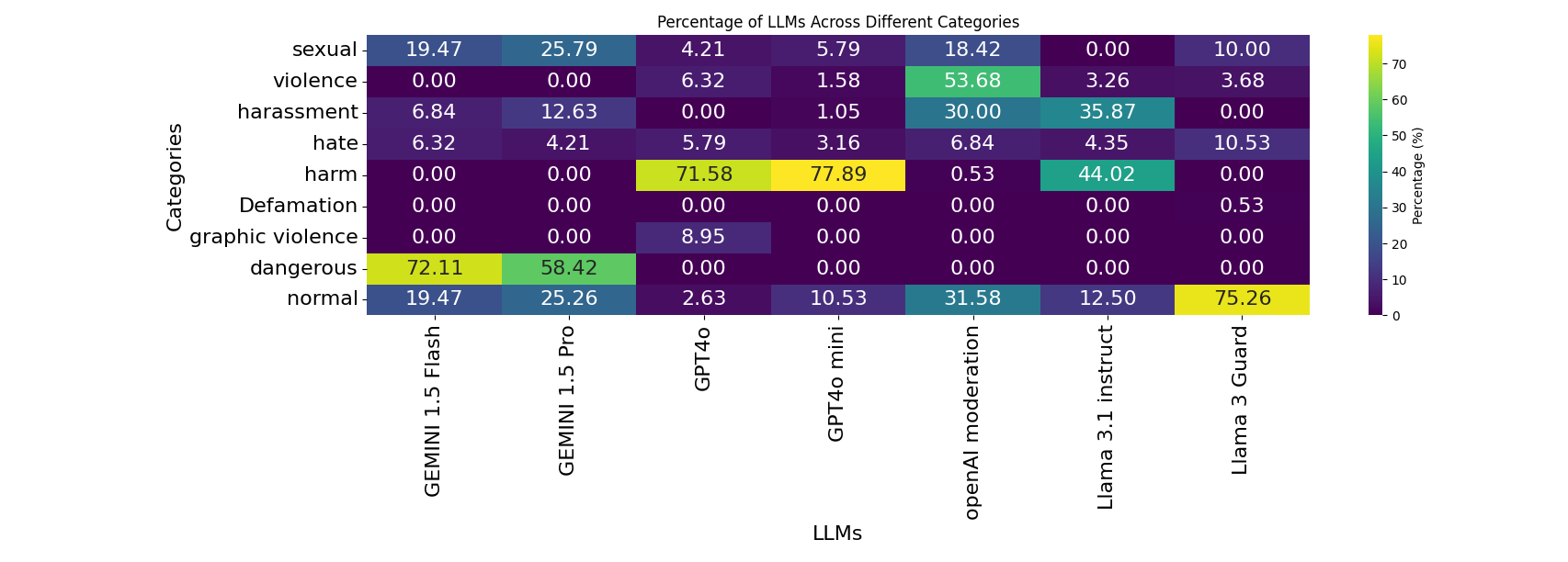}
    \caption{Heatmap of harmful practice showing predictions of various LLMs.}
    \label{fig:harmful_practice}
\end{figure}

Finally, the last category that requires to be assessed in tweets is the `harmful practices', shown in Figure~\ref{fig:harmful_practice}. This experiment indicates the limitations of \openaimod in detecting these harmful practices, giving the highest false negative errors (31.58\%). On the other hand, both \gpt and \gptmini have the lowest false negative errors, with \gpt leading by achieving the lowest error.

It is worth noting that \llamatext assigns the categories of `harassment', `hate', and `harm' to tweets, ignoring the `violence' and `sexual' categories in the five violence types. Additionally, \llamatext shows low false negative errors in all five types of violence.

In this experiment, the `harassment' category has been predicted by several LLMs. This reflects  the nuances in language that the models interpret based on the context and the wording. Overall, LLMs may not fully differentiate between the specific intents and contexts that separate `harassment' from the more specific terms like `sexual violence', `physical violence', `emotional violence', `economic violence', and `harmful practice'.

\subsubsection{Detection of inappropriate Content in Amazon Beauty Review}

In this experiment, we used the Amazon beauty reviews dataset~\cite{hou2024bridging} mentioned in the data overview section to present a use case of Amazon product reviews, aiming to study the use of LLMs for review censorship. Seven LLMs driven by the prompt given above have been examined to process the reviews. 

\begin{figure}[!htb]
    \centering
    \includegraphics[width=1\linewidth]{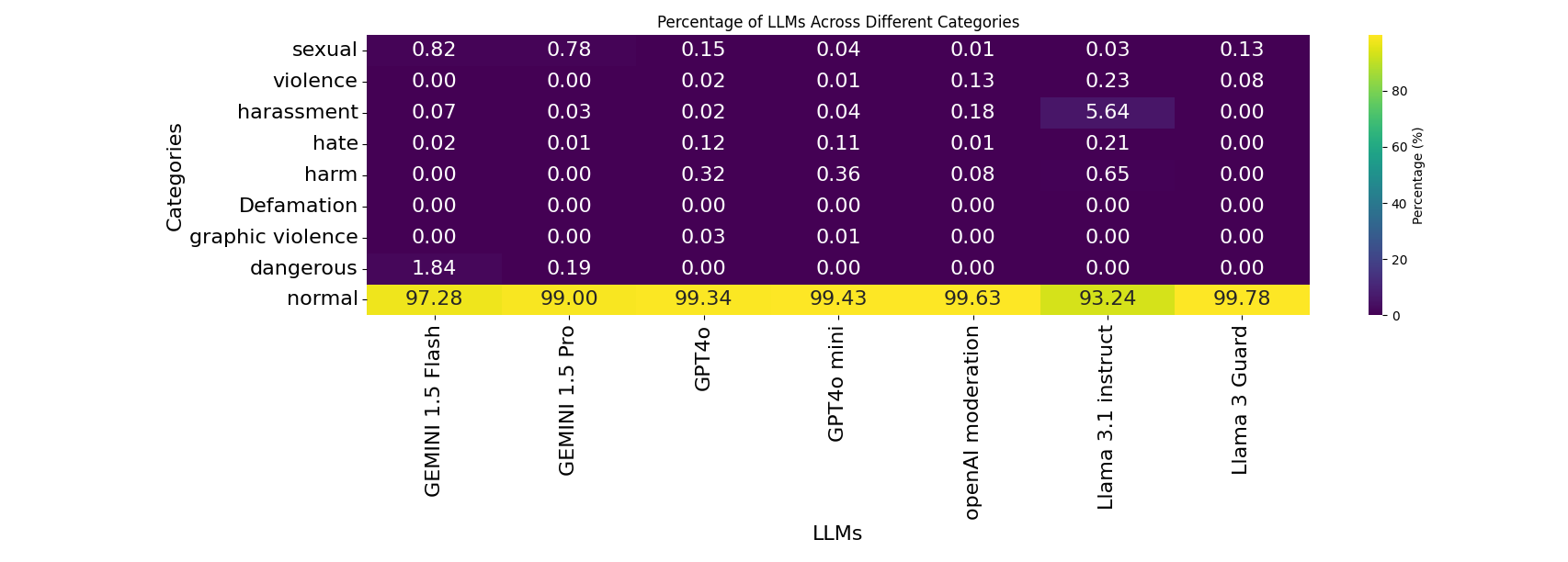}
    \caption{Heatmap of inappropriate content distribution in the Amazon beauty reviews.}
    \label{fig:review}
\end{figure}

The heatmap in Figure~\ref{fig:review} displays the percentage of several categories of content predicted by different LLMs.  All LLMs demonstrate similar performance in detecting normal contents when no inappropriate content is available. The majority of cells have low values, meaning that most categories do not have a significant presence of inappropriate content in these reviews. The heatmap suggests that most Amazon beauty reviews do not contain overtly inappropriate content, providing an answer to \textbf{RQ4}. 

Our finding aligns with Amazon's policy of making its reviews comply with Amazon Community Guidelines on appropriate language, content, videos, and images~\cite{Amazon_REcognizer_Content_Moderation}. This prevents reviews containing inappropriate language, offensive imagery, or any form of hate speech directed at individuals or communities from being displayed on the platform.

\subsubsection{Detection of inappropriate Content in News Articles}

In this experiment, we utilized the news article dataset outlined in the data overview section to demonstrate an additional use case. The main goal is to detect inappropriate content in news articles using seven LLMs guided by the prompt provided above. The whole article was fed to the inputs of LLMs for detection purposes.

\begin{figure}[!htb]
    \centering
    \includegraphics[width=1\linewidth]{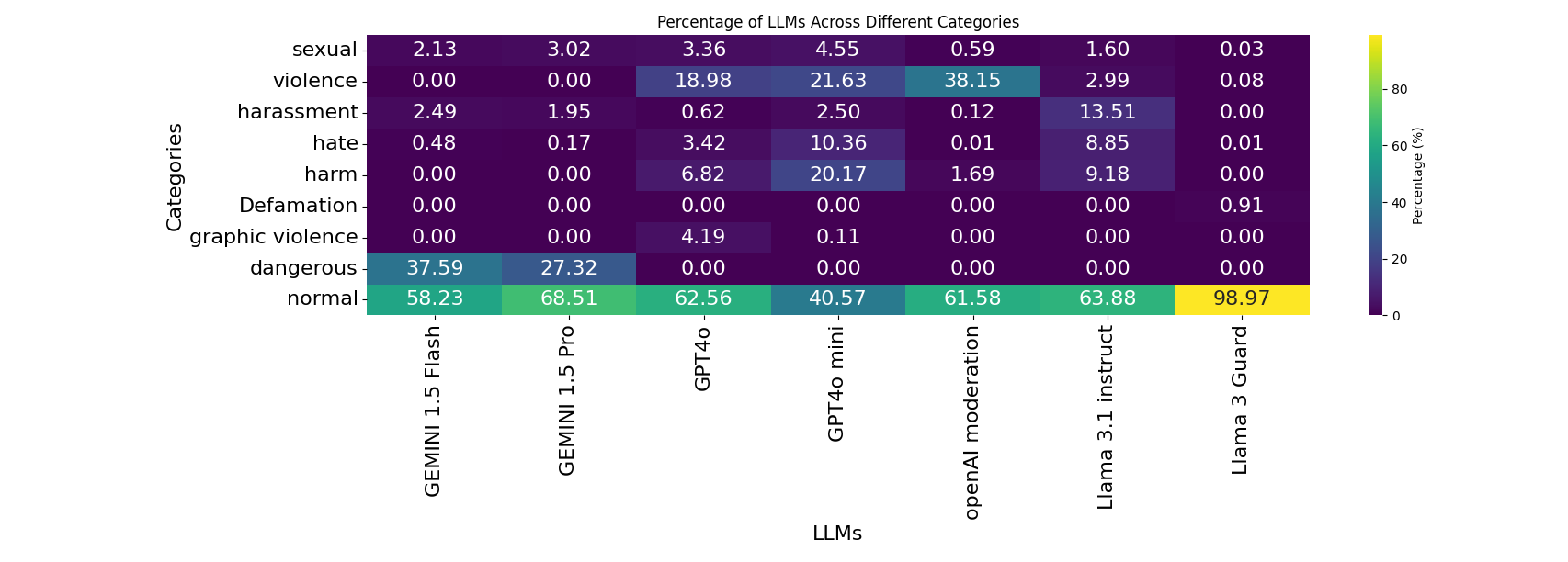}
    \caption{Heatmap of inappropriate content distribution in the CNN articles.}
    \label{fig:cnn}
\end{figure}

\begin{figure}[!htb]
    \centering
    \includegraphics[width=1\linewidth]{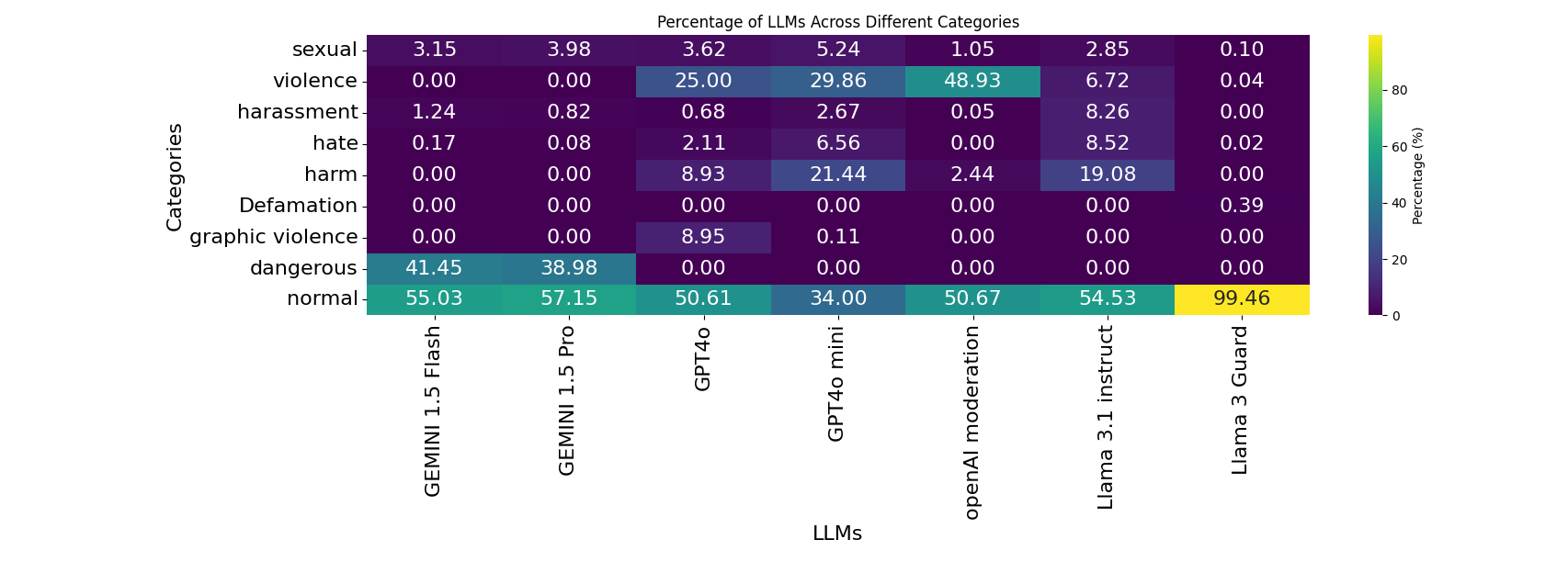}
    \caption{Heatmap of inappropriate content distribution in the Fox News articles.}
    \label{fig:foxnews}
\end{figure}

\begin{figure}[!htb]
    \centering
    \includegraphics[width=1\linewidth]{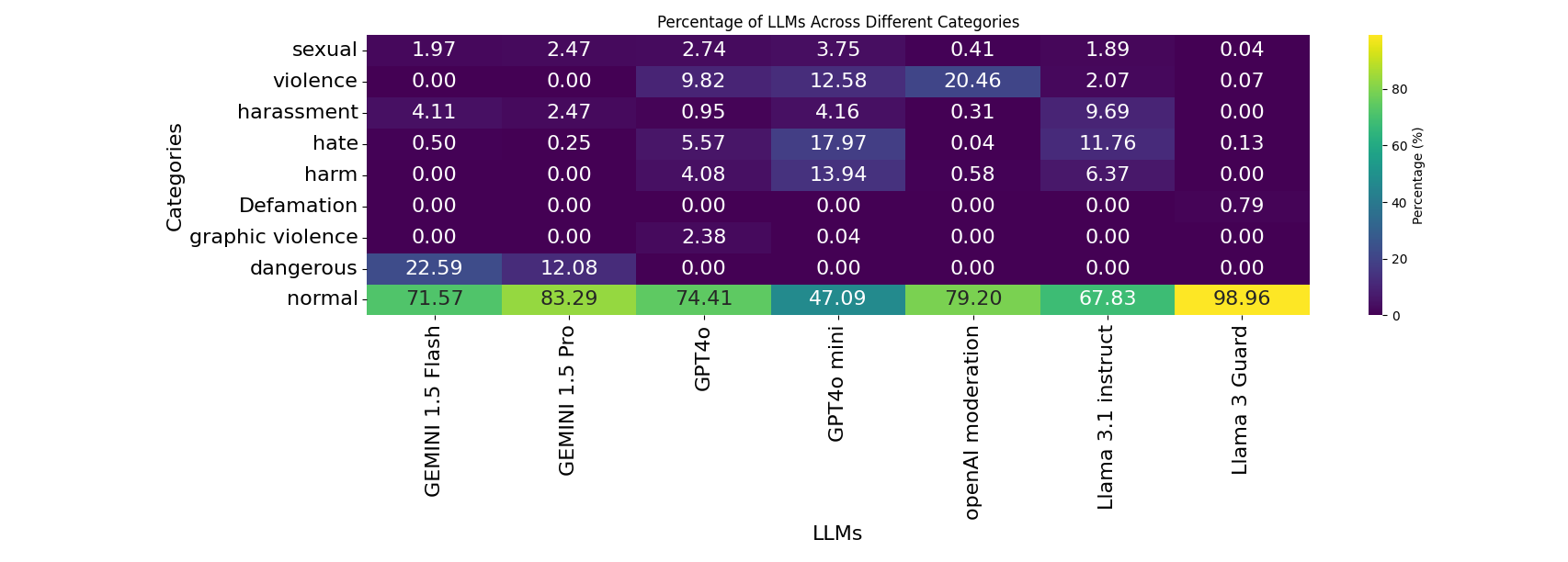}
    \caption{Heatmap of inappropriate content distribution in the Newsweek articles.}
    \label{fig:Newsweek}
\end{figure}

\begin{figure}[!htb]
    \centering
    \includegraphics[width=1\linewidth]{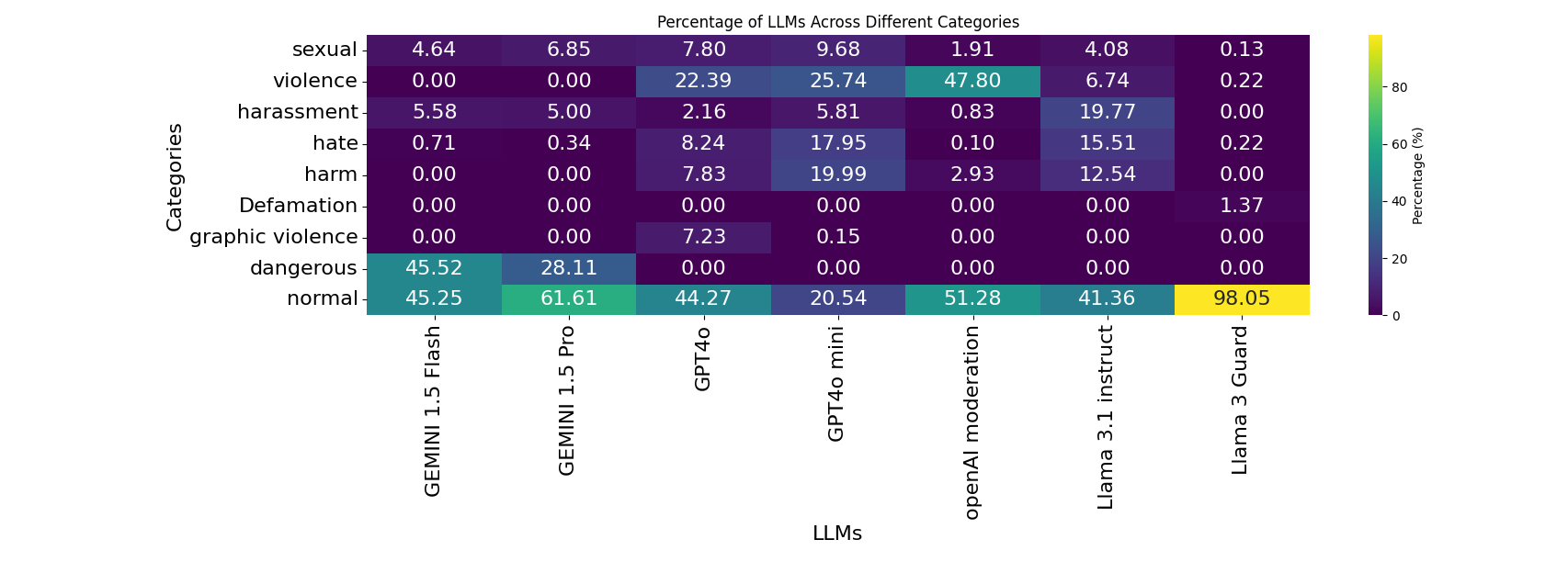}
    \caption{Heatmap of inappropriate content distribution in the Daily Beast articles.}
    \label{fig:DailyBeast}
\end{figure}

\begin{figure}[!htb]
    \centering
    \includegraphics[width=1\linewidth]{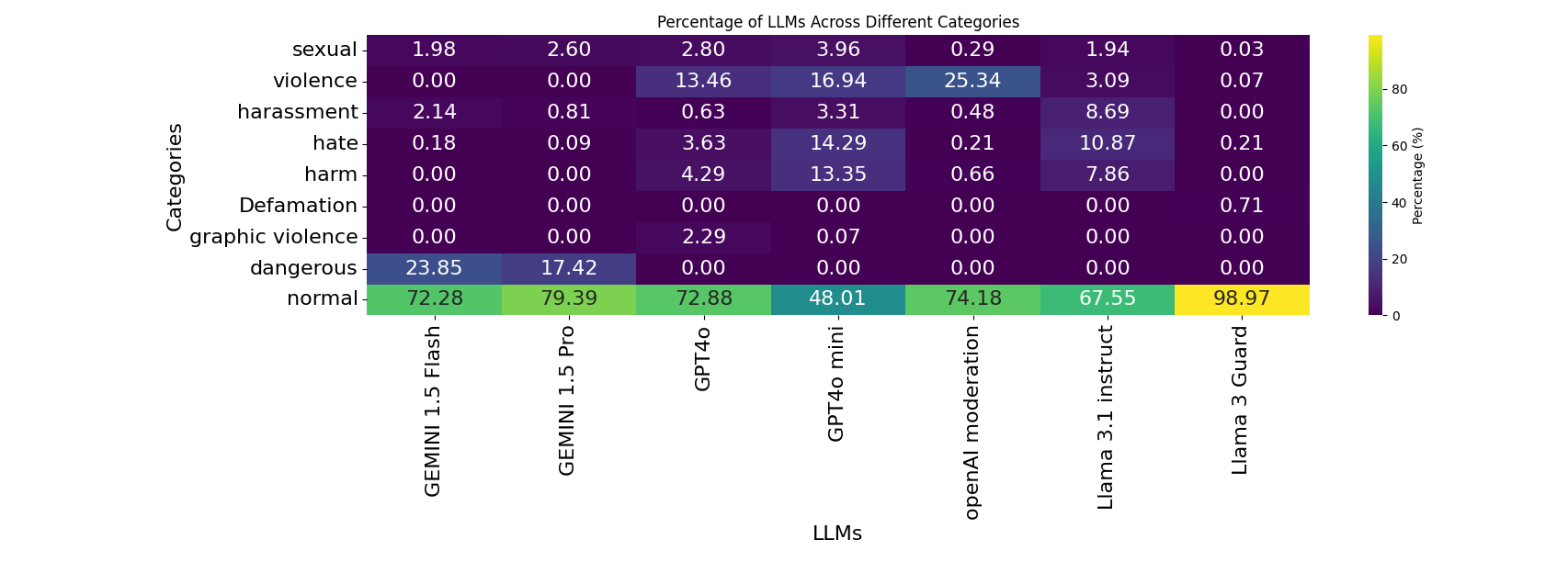}
    \caption{Heatmap of inappropriate content distribution in the Washington Times articles.}
    \label{fig:WashingtonTimes}
\end{figure}

The heatmaps, shown in~\Cref{fig:cnn,fig:foxnews,fig:Newsweek,fig:DailyBeast,fig:WashingtonTimes}, visualize the percentage distribution of responses across different categories for several LLMs. For nearly all models, the `normal' category has the highest percentage. This suggests that these models tend to classify a substantial portion of news content as safe. 

Due to fact saying that \llamaguardtext was primarily trained on conversational text, it may lack exposure to the diverse syntax and structure found in articles, making it less effective in making accurate predictions, which answers \textbf{RQ1} as shown in~\Cref{fig:cnn,fig:foxnews,fig:Newsweek,fig:DailyBeast,fig:WashingtonTimes}. As such, we exclude \llamaguardtext from the analysis below. In our future work, we would like to  consider segmenting the articles into smaller, coherent sections before processing them with \llamaguardtext which can help to manage the context more effectively.

In the CNN articles, \openaimod flags 38.15\% of the content as `violence'. \geminiflash marks 37.59\% as `dangerous (including violence and harm), trailed by \geminipro. Additionally, \gptmini categorizes 20.17\% of the content to fall under `harm'. Both \gptmini and \llamatext flagged content under the `hate' category, with \gptmini also showing the least content labeled as `normal'. This high sensitivity suggests this model may have stricter definitions of violence or danger in news articles.

For the Fox News articles, similar trends are observed: \openaimod flags 48.93\% as `violence', followed by \geminiflash (41.45\%), \geminipro (38.98\%), \gptmini (29.86\%), and \gpt (25\%). \gptmini also detects `harm' and `hate' content along with \llamatext. Again, \gptmini shows the lowest percentage in the `normal' category, indicating its high sensitivity.

Turning to Newsweek, \geminiflash flags 22.59\% of content as `violence', followed by \openaimod at 20.46\%, \gptmini at 12.58\%, \geminipro at 12.08\%, and \gpt at 9.82\%. \gptmini also identifies content in the `harm' and `hate' categories, along with \llamatext. Once again, \gptmini registers the lowest percentage in the `normal' category, reflecting its high sensitivity.

In the Daily Beast articles, similar patterns emerge. \openaimod flags 47.80\% of articles as `violence', followed by \geminiflash at 45.52\%, \geminipro at 28.11\%, \gptmini at 25.74\%, and \gpt at 22.39\%. In the `harm' category, both \gptmini and \llamatext detect 19.99\% and 12.54\% of articles, respectively, with both \gemini versions possibly categorizing these under the `dangerous' category. Additionally, both \gptmini and 
\llamatext flag `hate' content, with \gptmini recording the lowest percentage in the `normal' category, highlighting its heightened sensitivity to inappropriate content detection.

Concluding the analysis with the Washington Times, similar patterns are evident, with the \openaimod detecting 'violence' at the highest rate (25.34\%). Furthermore, \gptmini records the lowest percentage (48.01\%) in the 'normal' category, underscoring its increased sensitivity to identifying inappropriate content.

We analyzed and discussed the results from five news outlets, presenting the exact percentage distribution of inappropriate content categories such as sexual, harassment, hate, harm, violence, and dangerous for each news outlet, thereby addressing \textbf{RQ4}.

The `harassment' category was primarily identified by \llamatext across all news outlets, while it simultaneously showed a low percentage for detecting the `violence' category.

Our findings indicate similar patterns in the detection of inappropriate content across all news outlets, ensuring consistent moderation by LLMs across diverse sources. However, inconsistencies remain in the prediction accuracy of LLMs, with some detecting `harassment' while others detecting `violence,' with varying rates of false positives and false negatives.

Among the news outlets analyzed, the Daily Beast stands out with the highest percentage of articles flagged for `violence' and `harm'. Fox News follows, with most LLMs detecting higher levels of `violence' and `harm' in their articles compared to those from CNN. In contrast, Newsweek and the Washington Times have the lowest levels of `violence' or `harm' detected by the majority of LLMs. Similarly, the categories of `harassment' and `hate' are flagged in significant proportions (19.77\% and 15.51\%) in Daily Beast articles by \llamatext. Additionally, \gpt and \geminipro detect more sexual content in Daily Beast compared to other news sources.

\begin{figure}[!htb]
    \centering
    \includegraphics[width=1\textwidth]{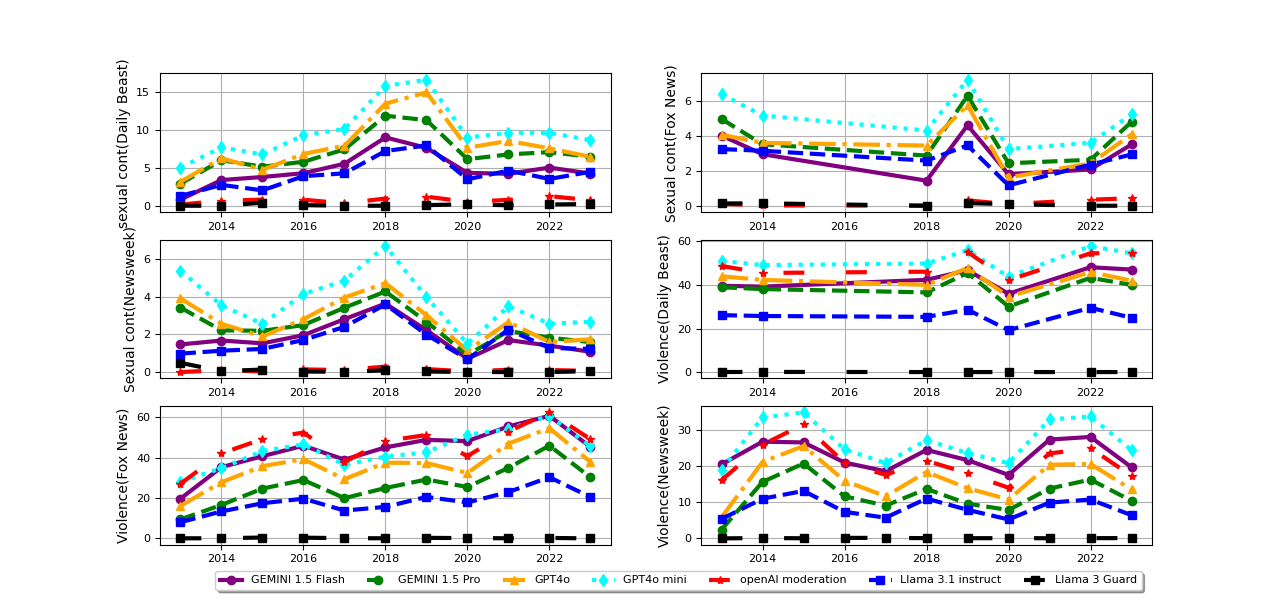}
    \caption{Trends of inappropriate content in media outlets: A comparative analysis of sexual and violent content detection across LLMs (2013-2023).}
    \label{fig:over_years}
\end{figure}

Figure~\ref{fig:over_years} compares the detection of sexual and violent content (samples predicted under `violence', `graphic violence', and `harm') in three major news outlets (Daily Beast, Newsweek, and Fox News) using different LLMs, showing trends over the years and highlighting variances in LLM performance. The figure shows that most LLMs follow similar patterns of increase and decrease in detecting inappropriate content over the years, even if their sensitivity levels vary. This alignment in trends suggests that, despite differences in detection thresholds, LLMs are responding to the underlying changes in the content across the media outlets.

For detecting sexual and violent content, \gptmini consistently identifies higher levels of such content compared to other models, indicating a greater sensitivity to these types of content. In contrast, \llamatext consistently detects lower levels of sexual and violent content relative to other models. Conversely, \llamaguardtext shows its inability to make accurate predictions. One reason behind this could be attributed to the fact that \llamaguardtext was primarily trained on conversational text. Therefore, it may lack familiarity with the diverse syntax and structure found in articles. 

\openaimod primarily targets clear violations of content guidelines. It is common for articles to contain implicit or nuanced content that might not be flagged by content moderation tools, such as \openaimod, as they are often calibrated for explicit material. In this experiment, \openaimod failed to detect sexual content in the three outlets tested. This may be due to differences between the datasets on which the model was trained on, and the articles used in this experiment. However, \openaimod demonstrated comparable performance to other LLMs in detecting violent content across these three outlets.

The Daily Beast chart shows higher levels of detected sexual content across all models, suggesting that Daily Beast articles may contain more sexual content. Conversely, Newsweek exhibits lower levels of detected violent content across all models, indicating that its articles may have fewer instances of violent material. 

We found that tasks of detecting violence, hate speech, and offensive language in tweets, adult content in articles' descriptions, and various categories of inappropriate contents in Amazon reviews and news articles answer \textbf{RQ3} by highlighting inconsistency in the prediction accuracy of the utilized LLMs.

\section{Conclusion and Future Work}
\label{sec:conclusion}

This paper demonstrated the challenging problem of textual and visual content censorship. We aimed to build a more efficient and accurate censorship system designed to reduce both false positives and false negatives. Therefore, we evaluated the large language models and studied their capabilities in identifying inappropriate contents across media outlets. To achieve this goal, various datasets like X tweets, Amazon reviews, articles from five news outlets, human photos, cartoons, sketches, and violence videos have
been utilized for evaluation and comparison. 

The heatmaps in the results and the discussion section illustrate that LLMs can differ significantly in interpreting the same content, providing an answer for \textbf{RQ3} and highlighting the importance of selecting the right model for specific content moderation needs.

The results demonstrate that LLMs outperform traditional techniques by achieving higher accuracy and lower false positive and false negative rates. This highlights their potential to be integrated into websites, social media platforms, and video-sharing services for regulatory and content moderation purposes.

Overall, we found that our proposed LLM-based content censorship system outperforms existing content moderation solutions such \openaimod and \llamaguard and highlights their limitations, answering \textbf{RQ1}. 

Our finding demonstrated \geminiflash as a good candidate LLM for visual content censorship, outperforming existing LLM-based content moderation such as \openaimod and \llamaguardvision, highlighting its ability to detect violence, nudity, pornography, and abuse in photos, cartoons, and sketches across both images and videos, providing an answer to \textbf{RQ2}. Its ability to balance high speed and high performance makes it effective for live-streaming platforms to respond and filter quickly and reliably. 

Focusing on textual content, we found that while the \openaimod is well-suited for detecting hate speech and offensive language, \gpt proves to be the optimal choice for identifying all forms of violence. Furthermore, both \gpt and \geminipro are excellent models for detecting adult content, outperforming existing LLM-based content moderation such as \openaimod and \llamaguardtext, thus providing an answer to \textbf{RQ2}.

Broadly speaking, in content moderation, a high FPR is generally preferred over a high false negative rate. The flagged content can be reviewed in a second stage by human moderators who confirm if the content is genuinely inappropriate. Prioritizing user safety means that it's less risky to over-flag potentially harmful content. The high FPR is manageable because it serves as a pre-filter, allowing human moderators to make the final call, thus ensuring both user safety and operational efficiency. 

In summary, content censorship for various media sources across diverse platforms—like news articles, online shopping reviews, social media posts and photos, YouTube comments, X tweets, and TikTok videos—requires sophisticated moderation techniques to monitor images, videos, and text. Each platform encounters unique challenges, but there's a common need to detect and manage content types. It is necessary to notify users with warnings or blur-sensitive material, advising viewers of the content risk and giving them control over their viewing choices.

For our future work, we plan to fine-tune LLMs to enhance their detection accuracy for news articles, particularly. However, labeling data remains challenging, as it requires human annotation of news articles involving violence, hate speech, and sexual content. This process demands subjective judgment, context awareness, and cultural sensitivity, often leading to inconsistencies among annotators. Crowd-sourcing platforms offer a faster, broader approach, allowing multiple annotators to label the same data point, helping to achieve consensus and reduce individual bias. Following this, any discrepancies can be flagged for expert review.

\newpage
\bibliographystyle{naturemag}
\bibliography{sample}

\begin{thebibliography}{100}
\expandafter\ifx\csname url\endcsname\relax
  \def\url#1{\texttt{#1}}\fi
\expandafter\ifx\csname urlprefix\endcsname\relax\def\urlprefix{URL }\fi
\providecommand{\bibinfo}[2]{#2}
\providecommand{\eprint}[2][]{\url{#2}}

\bibitem{salter2013justice}
\bibinfo{author}{Salter, M.}
\newblock \bibinfo{title}{Justice and revenge in online counter-publics: Emerging responses to sexual violence in the age of social media}.
\newblock \emph{\bibinfo{journal}{Crime, Media, Culture}} \textbf{\bibinfo{volume}{9}}, \bibinfo{pages}{225--242} (\bibinfo{year}{2013}).

\bibitem{strasburger2010health}
\bibinfo{author}{Strasburger, V.~C.}, \bibinfo{author}{Jordan, A.~B.} \& \bibinfo{author}{Donnerstein, E.}
\newblock \bibinfo{title}{Health effects of media on children and adolescents}.
\newblock \emph{\bibinfo{journal}{Pediatrics}} \textbf{\bibinfo{volume}{125}}, \bibinfo{pages}{756--767} (\bibinfo{year}{2010}).

\bibitem{greenfield2004inadvertent}
\bibinfo{author}{Greenfield, P.~M.}
\newblock \bibinfo{title}{Inadvertent exposure to pornography on the internet: Implications of peer-to-peer file-sharing networks for child development and families}.
\newblock \emph{\bibinfo{journal}{Journal of Applied Developmental Psychology}} \textbf{\bibinfo{volume}{25}}, \bibinfo{pages}{741--750} (\bibinfo{year}{2004}).

\bibitem{citron2018sexual}
\bibinfo{author}{Citron, D.~K.}
\newblock \bibinfo{title}{Sexual privacy}.
\newblock \emph{\bibinfo{journal}{Yale LJ}} \textbf{\bibinfo{volume}{128}}, \bibinfo{pages}{1870} (\bibinfo{year}{2018}).

\bibitem{krahe2011desensitization}
\bibinfo{author}{Krah{\'e}, B.} \emph{et~al.}
\newblock \bibinfo{title}{Desensitization to media violence: links with habitual media violence exposure, aggressive cognitions, and aggressive behavior.}
\newblock \emph{\bibinfo{journal}{Journal of personality and social psychology}} \textbf{\bibinfo{volume}{100}}, \bibinfo{pages}{630} (\bibinfo{year}{2011}).

\bibitem{Child_Pornography}
\bibinfo{title}{Child pornography}.
\newblock \bibinfo{howpublished}{\url{https://www.justice.gov/criminal/criminal-ceos/child-pornography}}.

\bibitem{powell2017sexual}
\bibinfo{author}{Powell, A.} \& \bibinfo{author}{Henry, N.}
\newblock \emph{\bibinfo{title}{Sexual violence in a digital age}} (\bibinfo{publisher}{Springer}, \bibinfo{year}{2017}).

\bibitem{Regulate_Tech}
\bibinfo{author}{Erin~Simpson, A.~C.}
\newblock \bibinfo{title}{How to regulate tech: A technology policy framework for online services}.
\newblock \bibinfo{howpublished}{\url{https://www.americanprogress.org/article/how-to-regulate-tech-a-technology-policy-framework-for-online-services}}.

\bibitem{impact_of_algorithms}
\bibinfo{author}{Giovanni~Sartor, A.~L., Giovanni~Sartor}.
\newblock \bibinfo{title}{The impact of algorithms for online content filtering or moderation}.
\newblock \bibinfo{howpublished}{\url{https://www.europarl.europa.eu/RegData/etudes/STUD/2020/657101/IPOL_STU(2020)657101_EN.pdf}}.

\bibitem{Freedom_of_Speech}
\bibinfo{title}{Freedom of speech in the digital age}.
\newblock \bibinfo{howpublished}{\url{https://newmajorityfoundation.com/freedom-of-speech/}}.

\bibitem{OpenAI_Moderation}
\bibinfo{author}{platform, O.}
\newblock \bibinfo{title}{Moderation}.
\newblock \bibinfo{howpublished}{\url{https://platform.openai.com/docs/guides/moderation}}.

\bibitem{markov2023holistic}
\bibinfo{author}{Markov, T.} \emph{et~al.}
\newblock \bibinfo{title}{A holistic approach to undesired content detection in the real world}.
\newblock In \emph{\bibinfo{booktitle}{Proceedings of the AAAI Conference on Artificial Intelligence}}, vol.~\bibinfo{volume}{37}, \bibinfo{pages}{15009--15018} (\bibinfo{year}{2023}).

\bibitem{inan2023llama}
\bibinfo{author}{Inan, H.} \emph{et~al.}
\newblock \bibinfo{title}{Llama guard: Llm-based input-output safeguard for human-ai conversations}.
\newblock \emph{\bibinfo{journal}{arXiv preprint arXiv:2312.06674}}  (\bibinfo{year}{2023}).

\bibitem{dubey2024llama3herdmodels}
\bibinfo{author}{Llama~Team, A. .~M.}
\newblock \bibinfo{title}{The llama 3 herd of models} (\bibinfo{year}{2024}).
\newblock \urlprefix\url{https://arxiv.org/abs/2407.21783}.
\newblock \eprint{2407.21783}.

\bibitem{aldahoulevaluation}
\bibinfo{author}{AlDahoul, N.}, \bibinfo{author}{Karim, H.~A.}, \bibinfo{author}{Momo, M.~A.}, \bibinfo{author}{Sy, M.~A.} \& \bibinfo{author}{Tan, M. J.~T.}
\newblock \bibinfo{title}{Evaluation of content moderation software for nudity and pornography detection in various scenarios}.
\newblock \emph{\bibinfo{journal}{MECON Multimedia University Engineering Conference}}  (\bibinfo{year}{2023}).

\bibitem{jahan2023systematic}
\bibinfo{author}{Jahan, M.~S.} \& \bibinfo{author}{Oussalah, M.}
\newblock \bibinfo{title}{A systematic review of hate speech automatic detection using natural language processing}.
\newblock \emph{\bibinfo{journal}{Neurocomputing}} \textbf{\bibinfo{volume}{546}}, \bibinfo{pages}{126232} (\bibinfo{year}{2023}).

\bibitem{dinakar2012common}
\bibinfo{author}{Dinakar, K.}, \bibinfo{author}{Jones, B.}, \bibinfo{author}{Havasi, C.}, \bibinfo{author}{Lieberman, H.} \& \bibinfo{author}{Picard, R.}
\newblock \bibinfo{title}{Common sense reasoning for detection, prevention, and mitigation of cyberbullying}.
\newblock \emph{\bibinfo{journal}{ACM Transactions on Interactive Intelligent Systems (TiiS)}} \textbf{\bibinfo{volume}{2}}, \bibinfo{pages}{1--30} (\bibinfo{year}{2012}).

\bibitem{abozinadah2016improved}
\bibinfo{author}{Abozinadah, E.~A.} \& \bibinfo{author}{Jones, J.~H.}
\newblock \bibinfo{title}{Improved micro-blog classification for detecting abusive arabic twitter accounts}.
\newblock \emph{\bibinfo{journal}{International Journal of Data Mining \& Knowledge Management Process (IJDKP)}} \textbf{\bibinfo{volume}{6}}, \bibinfo{pages}{17--28} (\bibinfo{year}{2016}).

\bibitem{badjatiya2017deep}
\bibinfo{author}{Badjatiya, P.}, \bibinfo{author}{Gupta, S.}, \bibinfo{author}{Gupta, M.} \& \bibinfo{author}{Varma, V.}
\newblock \bibinfo{title}{Deep learning for hate speech detection in tweets}.
\newblock In \emph{\bibinfo{booktitle}{Proceedings of the 26th international conference on World Wide Web companion}}, \bibinfo{pages}{759--760} (\bibinfo{year}{2017}).

\bibitem{davidson2017automated}
\bibinfo{author}{Davidson, T.}, \bibinfo{author}{Warmsley, D.}, \bibinfo{author}{Macy, M.} \& \bibinfo{author}{Weber, I.}
\newblock \bibinfo{title}{Automated hate speech detection and the problem of offensive language}.
\newblock In \emph{\bibinfo{booktitle}{Proceedings of the international AAAI conference on web and social media}}, vol.~\bibinfo{volume}{11}, \bibinfo{pages}{512--515} (\bibinfo{year}{2017}).

\bibitem{pawar2018cyberbullying}
\bibinfo{author}{Pawar, R.}, \bibinfo{author}{Agrawal, Y.}, \bibinfo{author}{Joshi, A.}, \bibinfo{author}{Gorrepati, R.} \& \bibinfo{author}{Raje, R.~R.}
\newblock \bibinfo{title}{Cyberbullying detection system with multiple server configurations}.
\newblock In \emph{\bibinfo{booktitle}{2018 IEEE International Conference on Electro/Information Technology (EIT)}}, \bibinfo{pages}{0090--0095} (\bibinfo{organization}{IEEE}, \bibinfo{year}{2018}).

\bibitem{ousidhoum2019multilingual}
\bibinfo{author}{Ousidhoum, N.}, \bibinfo{author}{Lin, Z.}, \bibinfo{author}{Zhang, H.}, \bibinfo{author}{Song, Y.} \& \bibinfo{author}{Yeung, D.-Y.}
\newblock \bibinfo{title}{Multilingual and multi-aspect hate speech analysis}.
\newblock \emph{\bibinfo{journal}{arXiv preprint arXiv:1908.11049}}  (\bibinfo{year}{2019}).

\bibitem{alakrot2018towards}
\bibinfo{author}{Alakrot, A.}, \bibinfo{author}{Murray, L.} \& \bibinfo{author}{Nikolov, N.~S.}
\newblock \bibinfo{title}{Towards accurate detection of offensive language in online communication in arabic}.
\newblock \emph{\bibinfo{journal}{Procedia computer science}} \textbf{\bibinfo{volume}{142}}, \bibinfo{pages}{315--320} (\bibinfo{year}{2018}).

\bibitem{malmasi2018challenges}
\bibinfo{author}{Malmasi, S.} \& \bibinfo{author}{Zampieri, M.}
\newblock \bibinfo{title}{Challenges in discriminating profanity from hate speech}.
\newblock \emph{\bibinfo{journal}{Journal of Experimental \& Theoretical Artificial Intelligence}} \textbf{\bibinfo{volume}{30}}, \bibinfo{pages}{187--202} (\bibinfo{year}{2018}).

\bibitem{kamble2018hate}
\bibinfo{author}{Kamble, S.} \& \bibinfo{author}{Joshi, A.}
\newblock \bibinfo{title}{Hate speech detection from code-mixed hindi-english tweets using deep learning models}.
\newblock \emph{\bibinfo{journal}{arXiv preprint arXiv:1811.05145}}  (\bibinfo{year}{2018}).

\bibitem{faris2020hate}
\bibinfo{author}{Faris, H.}, \bibinfo{author}{Aljarah, I.}, \bibinfo{author}{Habib, M.} \& \bibinfo{author}{Castillo, P.~A.}
\newblock \bibinfo{title}{Hate speech detection using word embedding and deep learning in the arabic language context.}
\newblock In \emph{\bibinfo{booktitle}{ICPRAM}}, \bibinfo{pages}{453--460} (\bibinfo{year}{2020}).

\bibitem{rizos2019augment}
\bibinfo{author}{Rizos, G.}, \bibinfo{author}{Hemker, K.} \& \bibinfo{author}{Schuller, B.}
\newblock \bibinfo{title}{Augment to prevent: short-text data augmentation in deep learning for hate-speech classification}.
\newblock In \emph{\bibinfo{booktitle}{Proceedings of the 28th ACM international conference on information and knowledge management}}, \bibinfo{pages}{991--1000} (\bibinfo{year}{2019}).

\bibitem{zhou2020deep}
\bibinfo{author}{Zhou, Y.}, \bibinfo{author}{Yang, Y.}, \bibinfo{author}{Liu, H.}, \bibinfo{author}{Liu, X.} \& \bibinfo{author}{Savage, N.}
\newblock \bibinfo{title}{Deep learning based fusion approach for hate speech detection}.
\newblock \emph{\bibinfo{journal}{IEEE Access}} \textbf{\bibinfo{volume}{8}}, \bibinfo{pages}{128923--128929} (\bibinfo{year}{2020}).

\bibitem{dowlagar2021hasocone}
\bibinfo{author}{Dowlagar, S.} \& \bibinfo{author}{Mamidi, R.}
\newblock \bibinfo{title}{Hasocone@ fire-hasoc2020: Using bert and multilingual bert models for hate speech detection}.
\newblock \emph{\bibinfo{journal}{arXiv preprint arXiv:2101.09007}}  (\bibinfo{year}{2021}).

\bibitem{mulki2019hsab}
\bibinfo{author}{Mulki, H.}, \bibinfo{author}{Haddad, H.}, \bibinfo{author}{Ali, C.~B.} \& \bibinfo{author}{Alshabani, H.}
\newblock \bibinfo{title}{L-hsab: A levantine twitter dataset for hate speech and abusive language}.
\newblock In \emph{\bibinfo{booktitle}{Proceedings of the third workshop on abusive language online}}, \bibinfo{pages}{111--118} (\bibinfo{year}{2019}).

\bibitem{yin2017comparative}
\bibinfo{author}{Yin, W.}, \bibinfo{author}{Kann, K.}, \bibinfo{author}{Yu, M.} \& \bibinfo{author}{Sch{\"u}tze, H.}
\newblock \bibinfo{title}{Comparative study of cnn and rnn for natural language processing}.
\newblock \emph{\bibinfo{journal}{arXiv preprint arXiv:1702.01923}}  (\bibinfo{year}{2017}).

\bibitem{pitsilis2018effective}
\bibinfo{author}{Pitsilis, G.~K.}, \bibinfo{author}{Ramampiaro, H.} \& \bibinfo{author}{Langseth, H.}
\newblock \bibinfo{title}{Effective hate-speech detection in twitter data using recurrent neural networks}.
\newblock \emph{\bibinfo{journal}{Applied Intelligence}} \textbf{\bibinfo{volume}{48}}, \bibinfo{pages}{4730--4742} (\bibinfo{year}{2018}).

\bibitem{alatawi2021detecting}
\bibinfo{author}{Alatawi, H.~S.}, \bibinfo{author}{Alhothali, A.~M.} \& \bibinfo{author}{Moria, K.~M.}
\newblock \bibinfo{title}{Detecting white supremacist hate speech using domain specific word embedding with deep learning and bert}.
\newblock \emph{\bibinfo{journal}{IEEE Access}} \textbf{\bibinfo{volume}{9}}, \bibinfo{pages}{106363--106374} (\bibinfo{year}{2021}).

\bibitem{ranasinghe2019brums}
\bibinfo{author}{Ranasinghe, T.}, \bibinfo{author}{Zampieri, M.} \& \bibinfo{author}{Hettiarachchi, H.}
\newblock \bibinfo{title}{Brums at hasoc 2019: Deep learning models for multilingual hate speech and offensive language identification.}
\newblock In \emph{\bibinfo{booktitle}{FIRE (working notes)}}, \bibinfo{pages}{199--207} (\bibinfo{year}{2019}).

\bibitem{polignano2019alberto}
\bibinfo{author}{Polignano, M.}, \bibinfo{author}{Basile, V.}, \bibinfo{author}{Basile, P.}, \bibinfo{author}{de~Gemmis, M.} \& \bibinfo{author}{Semeraro, G.}
\newblock \bibinfo{title}{Alberto: Modeling italian social media language with bert}.
\newblock \emph{\bibinfo{journal}{IJCoL. Italian Journal of Computational Linguistics}} \textbf{\bibinfo{volume}{5}}, \bibinfo{pages}{11--31} (\bibinfo{year}{2019}).

\bibitem{das2023team}
\bibinfo{author}{Das, R.} \emph{et~al.}
\newblock \bibinfo{title}{Team error point at blp-2023 task 1: A comprehensive approach for violence inciting text detection using deep learning and traditional machine learning algorithm}.
\newblock In \emph{\bibinfo{booktitle}{Proceedings of the First Workshop on Bangla Language Processing (BLP-2023)}}, \bibinfo{pages}{236--240} (\bibinfo{year}{2023}).

\bibitem{page2023mavericks}
\bibinfo{author}{Page, S.}, \bibinfo{author}{Mangalvedhekar, S.}, \bibinfo{author}{Deshpande, K.}, \bibinfo{author}{Chavan, T.} \& \bibinfo{author}{Sonawane, S.}
\newblock \bibinfo{title}{Mavericks at blp-2023 task 1: Ensemble-based approach using language models for violence inciting text detection}.
\newblock In \emph{\bibinfo{booktitle}{Proceedings of the first workshop on bangla language processing (BLP-2023)}}, \bibinfo{pages}{190--195} (\bibinfo{year}{2023}).

\bibitem{khan2024detection}
\bibinfo{author}{Khan, M.~S.}, \bibinfo{author}{Malik, M. S.~I.} \& \bibinfo{author}{Nadeem, A.}
\newblock \bibinfo{title}{Detection of violence incitation expressions in urdu tweets using convolutional neural network}.
\newblock \emph{\bibinfo{journal}{Expert Systems with Applications}} \textbf{\bibinfo{volume}{245}}, \bibinfo{pages}{123174} (\bibinfo{year}{2024}).

\bibitem{ba2021design}
\bibinfo{author}{Ba~Wazir, A.~S.} \emph{et~al.}
\newblock \bibinfo{title}{Design and implementation of fast spoken foul language recognition with different end-to-end deep neural network architectures}.
\newblock \emph{\bibinfo{journal}{Sensors}} \textbf{\bibinfo{volume}{21}}, \bibinfo{pages}{710} (\bibinfo{year}{2021}).

\bibitem{wazir2020spectrogram}
\bibinfo{author}{Wazir, A. S.~B.} \emph{et~al.}
\newblock \bibinfo{title}{Spectrogram-based classification of spoken foul language using deep cnn}.
\newblock In \emph{\bibinfo{booktitle}{2020 IEEE 22nd International Workshop on Multimedia Signal Processing (MMSP)}}, \bibinfo{pages}{1--6} (\bibinfo{organization}{IEEE}, \bibinfo{year}{2020}).

\bibitem{nguyen2023fine}
\bibinfo{author}{Nguyen, T.~T.}, \bibinfo{author}{Wilson, C.} \& \bibinfo{author}{Dalins, J.}
\newblock \bibinfo{title}{Fine-tuning llama 2 large language models for detecting online sexual predatory chats and abusive texts}.
\newblock \emph{\bibinfo{journal}{arXiv preprint arXiv:2308.14683}}  (\bibinfo{year}{2023}).

\bibitem{vogt2021early}
\bibinfo{author}{Vogt, M.}, \bibinfo{author}{Leser, U.} \& \bibinfo{author}{Akbik, A.}
\newblock \bibinfo{title}{Early detection of sexual predators in chats}.
\newblock In \emph{\bibinfo{booktitle}{Proceedings of the 59th Annual Meeting of the Association for Computational Linguistics and the 11th International Joint Conference on Natural Language Processing (Volume 1: Long Papers)}}, \bibinfo{pages}{4985--4999} (\bibinfo{year}{2021}).

\bibitem{hamzah2021detection}
\bibinfo{author}{Hamzah, N.~A.} \& \bibinfo{author}{Dhannoon, B.~N.}
\newblock \bibinfo{title}{The detection of sexual harassment and chat predators using artificial neural network}.
\newblock \emph{\bibinfo{journal}{Karbala International Journal of Modern Science}} \textbf{\bibinfo{volume}{7}}, \bibinfo{pages}{6} (\bibinfo{year}{2021}).

\bibitem{yan2021bert}
\bibinfo{author}{Yan, M.} \& \bibinfo{author}{Luo, X.}
\newblock \bibinfo{title}{Bert-based detection of sexual harassment in dialogues}.
\newblock In \emph{\bibinfo{booktitle}{Proceedings of the 2021 5th International Conference on Computer Science and Artificial Intelligence}}, \bibinfo{pages}{359--364} (\bibinfo{year}{2021}).

\bibitem{ketsbaia2020detection}
\bibinfo{author}{Ketsbaia, L.}, \bibinfo{author}{Issac, B.} \& \bibinfo{author}{Chen, X.}
\newblock \bibinfo{title}{Detection of hate tweets using machine learning and deep learning}.
\newblock In \emph{\bibinfo{booktitle}{2020 IEEE 19th International Conference on Trust, Security and Privacy in Computing and Communications (TrustCom)}}, \bibinfo{pages}{751--758} (\bibinfo{organization}{IEEE}, \bibinfo{year}{2020}).

\bibitem{khezzar2023arhatedetector}
\bibinfo{author}{Khezzar, R.}, \bibinfo{author}{Moursi, A.} \& \bibinfo{author}{Al~Aghbari, Z.}
\newblock \bibinfo{title}{arhatedetector: detection of hate speech from standard and dialectal arabic tweets}.
\newblock \emph{\bibinfo{journal}{Discover Internet of Things}} \textbf{\bibinfo{volume}{3}}, \bibinfo{pages}{1} (\bibinfo{year}{2023}).

\bibitem{al2022detection}
\bibinfo{author}{Al-Hassan, A.} \& \bibinfo{author}{Al-Dossari, H.}
\newblock \bibinfo{title}{Detection of hate speech in arabic tweets using deep learning}.
\newblock \emph{\bibinfo{journal}{Multimedia systems}} \textbf{\bibinfo{volume}{28}}, \bibinfo{pages}{1963--1974} (\bibinfo{year}{2022}).

\bibitem{guelorget2021active}
\bibinfo{author}{Gu{\'e}lorget, P.}, \bibinfo{author}{Gadek, G.}, \bibinfo{author}{Zaharia, T.} \& \bibinfo{author}{Grilheres, B.}
\newblock \bibinfo{title}{Active learning to measure opinion and violence in french newspapers}.
\newblock \emph{\bibinfo{journal}{Procedia Computer Science}} \textbf{\bibinfo{volume}{192}}, \bibinfo{pages}{202--211} (\bibinfo{year}{2021}).

\bibitem{bello2020machine}
\bibinfo{author}{Bello, H.~J.}, \bibinfo{author}{Palomar, N.}, \bibinfo{author}{Gallego, E.}, \bibinfo{author}{Navascu{\'e}s, L.~J.} \& \bibinfo{author}{Lozano, C.}
\newblock \bibinfo{title}{Machine learning to study the impact of gender-based violence in the news media}.
\newblock \emph{\bibinfo{journal}{arXiv preprint arXiv:2012.07490}}  (\bibinfo{year}{2020}).

\bibitem{del2017hate}
\bibinfo{author}{Del~Vigna12, F.}, \bibinfo{author}{Cimino23, A.}, \bibinfo{author}{Dell’Orletta, F.}, \bibinfo{author}{Petrocchi, M.} \& \bibinfo{author}{Tesconi, M.}
\newblock \bibinfo{title}{Hate me, hate me not: Hate speech detection on facebook}.
\newblock In \emph{\bibinfo{booktitle}{Proceedings of the first Italian conference on cybersecurity (ITASEC17)}}, \bibinfo{pages}{86--95} (\bibinfo{year}{2017}).

\bibitem{palazon2023identifying}
\bibinfo{author}{Palaz{\'o}n-Fern{\'a}ndez, J.~L.} \emph{et~al.}
\newblock \bibinfo{title}{Identifying hate speech and attribution of responsibility: An analysis of simulated whatsapp conversations during the pandemic}.
\newblock In \emph{\bibinfo{booktitle}{Healthcare}}, vol.~\bibinfo{volume}{11}, \bibinfo{pages}{1564} (\bibinfo{organization}{MDPI}, \bibinfo{year}{2023}).

\bibitem{ries2014survey}
\bibinfo{author}{Ries, C.~X.} \& \bibinfo{author}{Lienhart, R.}
\newblock \bibinfo{title}{A survey on visual adult image recognition}.
\newblock \emph{\bibinfo{journal}{Multimedia tools and applications}} \textbf{\bibinfo{volume}{69}}, \bibinfo{pages}{661--688} (\bibinfo{year}{2014}).

\bibitem{de2012statistical}
\bibinfo{author}{de~Castro~Polastro, M.} \& \bibinfo{author}{da~Silva~Eleuterio, P.~M.}
\newblock \bibinfo{title}{A statistical approach for identifying videos of child pornography at crime scenes}.
\newblock In \emph{\bibinfo{booktitle}{2012 Seventh International Conference on Availability, Reliability and Security}}, \bibinfo{pages}{604--612} (\bibinfo{organization}{IEEE}, \bibinfo{year}{2012}).

\bibitem{arentz2004classifying}
\bibinfo{author}{Arentz, W.~A.} \& \bibinfo{author}{Olstad, B.}
\newblock \bibinfo{title}{Classifying offensive sites based on image content}.
\newblock \emph{\bibinfo{journal}{Computer Vision and Image Understanding}} \textbf{\bibinfo{volume}{94}}, \bibinfo{pages}{295--310} (\bibinfo{year}{2004}).

\bibitem{zaidan2015robust}
\bibinfo{author}{Zaidan, A.}, \bibinfo{author}{Karim, H.~A.}, \bibinfo{author}{Ahmad, N.~N.}, \bibinfo{author}{Zaidan, B.~B.} \& \bibinfo{author}{Kiah, M.~M.}
\newblock \bibinfo{title}{Robust pornography classification solving the image size variation problem based on multi-agent learning}.
\newblock \emph{\bibinfo{journal}{Journal of Circuits, Systems and Computers}} \textbf{\bibinfo{volume}{24}}, \bibinfo{pages}{1550023} (\bibinfo{year}{2015}).

\bibitem{wang1997system}
\bibinfo{author}{Wang, J.~Z.}, \bibinfo{author}{Wiederhold, G.} \& \bibinfo{author}{Firschein, O.}
\newblock \bibinfo{title}{System for screening objectionable images using daubechies' wavelets and color histograms}.
\newblock In \emph{\bibinfo{booktitle}{Interactive Distributed Multimedia Systems and Telecommunication Services: 4th International Workshop, IDMS'97 Darmstadt, Germany, September 10--12, 1997 Proceedings 4}}, \bibinfo{pages}{20--30} (\bibinfo{organization}{Springer}, \bibinfo{year}{1997}).

\bibitem{zheng2006shape}
\bibinfo{author}{Zheng, Q.-F.}, \bibinfo{author}{Zeng, W.}, \bibinfo{author}{Wang, W.-Q.} \& \bibinfo{author}{Gao, W.}
\newblock \bibinfo{title}{Shape-based adult image detection}.
\newblock \emph{\bibinfo{journal}{International Journal of Image and Graphics}} \textbf{\bibinfo{volume}{6}}, \bibinfo{pages}{115--124} (\bibinfo{year}{2006}).

\bibitem{fleck1996finding}
\bibinfo{author}{Fleck, M.~M.}, \bibinfo{author}{Forsyth, D.~A.} \& \bibinfo{author}{Bregler, C.}
\newblock \bibinfo{title}{Finding naked people}.
\newblock In \emph{\bibinfo{booktitle}{Computer Vision—ECCV'96: 4th European Conference on Computer Vision Cambridge, UK, April 15--18, 1996 Proceedings Volume II 4}}, \bibinfo{pages}{593--602} (\bibinfo{organization}{Springer}, \bibinfo{year}{1996}).

\bibitem{bosson2002non}
\bibinfo{author}{Bosson, A.}, \bibinfo{author}{Cawley, G.~C.}, \bibinfo{author}{Chan, Y.} \& \bibinfo{author}{Harvey, R.}
\newblock \bibinfo{title}{Non-retrieval: blocking pornographic images}.
\newblock In \emph{\bibinfo{booktitle}{International Conference on Image and Video Retrieval}}, \bibinfo{pages}{50--60} (\bibinfo{organization}{Springer}, \bibinfo{year}{2002}).

\bibitem{kim2005detecting}
\bibinfo{author}{Kim, W.}, \bibinfo{author}{Yoo, S.~J.}, \bibinfo{author}{Kim, J.-s.}, \bibinfo{author}{Nam, T.~Y.} \& \bibinfo{author}{Yoon, K.}
\newblock \bibinfo{title}{Detecting adult images using seven mpeg-7 visual descriptors}.
\newblock In \emph{\bibinfo{booktitle}{Web and Communication Technologies and Internet-Related Social Issues-HSI 2005: 3rd International Conference on Human. Society@ Internet, Tokyo, Japan, July 27-29, 2005. Proceedings 3}}, \bibinfo{pages}{336--339} (\bibinfo{organization}{Springer}, \bibinfo{year}{2005}).

\bibitem{wijaya2015phonographic}
\bibinfo{author}{Wijaya, I. G. P.~S.}, \bibinfo{author}{Widiartha, I.}, \bibinfo{author}{Uchimura, K.} \& \bibinfo{author}{Koutaki, G.}
\newblock \bibinfo{title}{Phonographic image recognition using fusion of scale invariant descriptor}.
\newblock In \emph{\bibinfo{booktitle}{2015 21st Korea-Japan Joint Workshop on Frontiers of Computer Vision (FCV)}}, \bibinfo{pages}{1--5} (\bibinfo{organization}{IEEE}, \bibinfo{year}{2015}).

\bibitem{lienhart2009filtering}
\bibinfo{author}{Lienhart, R.} \& \bibinfo{author}{Hauke, R.}
\newblock \bibinfo{title}{Filtering adult image content with topic models}.
\newblock In \emph{\bibinfo{booktitle}{2009 IEEE International Conference on Multimedia and Expo}}, \bibinfo{pages}{1472--1475} (\bibinfo{organization}{IEEE}, \bibinfo{year}{2009}).

\bibitem{caetano2014pornography}
\bibinfo{author}{Caetano, C.}, \bibinfo{author}{Avila, S.}, \bibinfo{author}{Guimaraes, S.} \& \bibinfo{author}{Ara{\'u}jo, A. d.~A.}
\newblock \bibinfo{title}{Pornography detection using bossanova video descriptor}.
\newblock In \emph{\bibinfo{booktitle}{2014 22nd European Signal Processing Conference (EUSIPCO)}}, \bibinfo{pages}{1681--1685} (\bibinfo{organization}{IEEE}, \bibinfo{year}{2014}).

\bibitem{caetano2016mid}
\bibinfo{author}{Caetano, C.}, \bibinfo{author}{Avila, S.}, \bibinfo{author}{Schwartz, W.~R.}, \bibinfo{author}{Guimar{\~a}es, S. J.~F.} \& \bibinfo{author}{Ara{\'u}jo, A. d.~A.}
\newblock \bibinfo{title}{A mid-level video representation based on binary descriptors: A case study for pornography detection}.
\newblock \emph{\bibinfo{journal}{Neurocomputing}} \textbf{\bibinfo{volume}{213}}, \bibinfo{pages}{102--114} (\bibinfo{year}{2016}).

\bibitem{jin2018pornographic}
\bibinfo{author}{Jin, X.}, \bibinfo{author}{Wang, Y.} \& \bibinfo{author}{Tan, X.}
\newblock \bibinfo{title}{Pornographic image recognition via weighted multiple instance learning}.
\newblock \emph{\bibinfo{journal}{IEEE transactions on cybernetics}} \textbf{\bibinfo{volume}{49}}, \bibinfo{pages}{4412--4420} (\bibinfo{year}{2018}).

\bibitem{lipornographic}
\bibinfo{author}{Li, D.}, \bibinfo{author}{Ji, Z.} \emph{et~al.}
\newblock \bibinfo{title}{Pornographic image recognition based on multi-instance deep learning}.
\newblock \emph{\bibinfo{journal}{SSRN}}  (\bibinfo{year}{2022}).

\bibitem{perez2017video}
\bibinfo{author}{Perez, M.} \emph{et~al.}
\newblock \bibinfo{title}{Video pornography detection through deep learning techniques and motion information}.
\newblock \emph{\bibinfo{journal}{Neurocomputing}} \textbf{\bibinfo{volume}{230}}, \bibinfo{pages}{279--293} (\bibinfo{year}{2017}).

\bibitem{moustafa2015applying}
\bibinfo{author}{Moustafa, M.}
\newblock \bibinfo{title}{Applying deep learning to classify pornographic images and videos}.
\newblock \emph{\bibinfo{journal}{arXiv preprint arXiv:1511.08899}}  (\bibinfo{year}{2015}).

\bibitem{hor2021evaluation}
\bibinfo{author}{Hor, S.~L.} \emph{et~al.}
\newblock \bibinfo{title}{An evaluation of state-of-the-art object detectors for pornography detection}.
\newblock In \emph{\bibinfo{booktitle}{2021 IEEE International Conference on Signal and Image Processing Applications (ICSIPA)}}, \bibinfo{pages}{191--196} (\bibinfo{organization}{IEEE}, \bibinfo{year}{2021}).

\bibitem{aldahoul2019local}
\bibinfo{author}{AlDahoul, N.} \emph{et~al.}
\newblock \bibinfo{title}{Local receptive field-extreme learning machine based adult content detection}.
\newblock In \emph{\bibinfo{booktitle}{2019 IEEE International Conference on Signal and Image Processing Applications (ICSIPA)}}, \bibinfo{pages}{128--133} (\bibinfo{organization}{IEEE}, \bibinfo{year}{2019}).

\bibitem{nian2016pornographic}
\bibinfo{author}{Nian, F.}, \bibinfo{author}{Li, T.}, \bibinfo{author}{Wang, Y.}, \bibinfo{author}{Xu, M.} \& \bibinfo{author}{Wu, J.}
\newblock \bibinfo{title}{Pornographic image detection utilizing deep convolutional neural networks}.
\newblock \emph{\bibinfo{journal}{Neurocomputing}} \textbf{\bibinfo{volume}{210}}, \bibinfo{pages}{283--293} (\bibinfo{year}{2016}).

\bibitem{aldahoul2020transfer}
\bibinfo{author}{AlDahoul, N.} \emph{et~al.}
\newblock \bibinfo{title}{Transfer detection of yolo to focus cnn’s attention on nude regions for adult content detection}.
\newblock \emph{\bibinfo{journal}{Symmetry}} \textbf{\bibinfo{volume}{13}}, \bibinfo{pages}{26} (\bibinfo{year}{2020}).

\bibitem{lyn2020convolutional}
\bibinfo{author}{Lyn, H.~S.}, \bibinfo{author}{Mansor, S.}, \bibinfo{author}{AlDahoul, N.} \& \bibinfo{author}{Karim, H.~A.}
\newblock \bibinfo{title}{Convolutional neural network-based transfer learning and classification of visual contents for film censorship}.
\newblock \emph{\bibinfo{journal}{Journal of Engineering Technology and Applied Physics}} \textbf{\bibinfo{volume}{2}}, \bibinfo{pages}{28--35} (\bibinfo{year}{2020}).

\bibitem{hor2022deep}
\bibinfo{author}{Hor, S.~L.} \emph{et~al.}
\newblock \bibinfo{title}{Deep active learning for pornography recognition using resnet}.
\newblock \emph{\bibinfo{journal}{International Journal of Technology}} \textbf{\bibinfo{volume}{13}}, \bibinfo{pages}{1261--1270} (\bibinfo{year}{2022}).

\bibitem{aldahoul2021evaluation}
\bibinfo{author}{Aldahoul, N.} \emph{et~al.}
\newblock \bibinfo{title}{An evaluation of traditional and cnn-based feature descriptors for cartoon pornography detection}.
\newblock \emph{\bibinfo{journal}{IEEE Access}} \textbf{\bibinfo{volume}{9}}, \bibinfo{pages}{39910--39925} (\bibinfo{year}{2021}).

\bibitem{aldahoul2021comparative}
\bibinfo{author}{AlDahoul, N.}, \bibinfo{author}{Karim, H.~A.}, \bibinfo{author}{Wazir, A. S.~B.}, \bibinfo{author}{Momo, M.~A.} \& \bibinfo{author}{Abdullah, M. H.~L.}
\newblock \bibinfo{title}{A comparative study of in-domain vs cross-domain learning for porn cartoon classification}.
\newblock In \emph{\bibinfo{booktitle}{2021 IEEE International Conference on Signal and Image Processing Applications (ICSIPA)}}, \bibinfo{pages}{60--65} (\bibinfo{organization}{IEEE}, \bibinfo{year}{2021}).

\bibitem{momo2023evaluation}
\bibinfo{author}{Momo, M.~A.} \emph{et~al.}
\newblock \bibinfo{title}{Evaluation of convolution and attention networks for nudity and pornography detection in sketch images}.
\newblock In \emph{\bibinfo{booktitle}{2023 IEEE Symposium on Computers \& Informatics (ISCI)}}, \bibinfo{pages}{7--12} (\bibinfo{organization}{IEEE}, \bibinfo{year}{2023}).

\bibitem{huang2016using}
\bibinfo{author}{Huang, Y.} \& \bibinfo{author}{Kong, A. W.~K.}
\newblock \bibinfo{title}{Using a cnn ensemble for detecting pornographic and upskirt images}.
\newblock In \emph{\bibinfo{booktitle}{2016 IEEE 8th International Conference on Biometrics Theory, Applications and Systems (BTAS)}}, \bibinfo{pages}{1--7} (\bibinfo{organization}{IEEE}, \bibinfo{year}{2016}).

\bibitem{wang2022pornographic}
\bibinfo{author}{Wang, Y.} \& \bibinfo{author}{Li, W.}
\newblock \bibinfo{title}{Pornographic image recognition based on high and low level feature fusion with human body masking and attention}.
\newblock In \emph{\bibinfo{booktitle}{Proceedings of the 2022 3rd International Conference on Control, Robotics and Intelligent System}}, \bibinfo{pages}{199--204} (\bibinfo{year}{2022}).

\bibitem{febin2020violence}
\bibinfo{author}{Febin, I.}, \bibinfo{author}{Jayasree, K.} \& \bibinfo{author}{Joy, P.~T.}
\newblock \bibinfo{title}{Violence detection in videos for an intelligent surveillance system using mobsift and movement filtering algorithm}.
\newblock \emph{\bibinfo{journal}{Pattern Analysis and Applications}} \textbf{\bibinfo{volume}{23}}, \bibinfo{pages}{611--623} (\bibinfo{year}{2020}).

\bibitem{nadeem2019wvd}
\bibinfo{author}{Nadeem, M.~S.}, \bibinfo{author}{Franqueira, V.~N.}, \bibinfo{author}{Kurugollu, F.} \& \bibinfo{author}{Zhai, X.}
\newblock \bibinfo{title}{Wvd: A new synthetic dataset for video-based violence detection}.
\newblock In \emph{\bibinfo{booktitle}{Artificial Intelligence XXXVI: 39th SGAI International Conference on Artificial Intelligence, AI 2019, Cambridge, UK, December 17--19, 2019, Proceedings 39}}, \bibinfo{pages}{158--164} (\bibinfo{organization}{Springer}, \bibinfo{year}{2019}).

\bibitem{wang2012baseline}
\bibinfo{author}{Wang, D.}, \bibinfo{author}{Zhang, Z.}, \bibinfo{author}{Wang, W.}, \bibinfo{author}{Wang, L.} \& \bibinfo{author}{Tan, T.}
\newblock \bibinfo{title}{Baseline results for violence detection in still images}.
\newblock In \emph{\bibinfo{booktitle}{2012 IEEE Ninth International Conference on Advanced Video and Signal-Based Surveillance}}, \bibinfo{pages}{54--57} (\bibinfo{organization}{IEEE}, \bibinfo{year}{2012}).

\bibitem{patel2021real}
\bibinfo{author}{Patel, M.}
\newblock \bibinfo{title}{Real-time violence detection using cnn-lstm}.
\newblock \emph{\bibinfo{journal}{arXiv preprint arXiv:2107.07578}}  (\bibinfo{year}{2021}).

\bibitem{ali2018violence}
\bibinfo{author}{Ali, A.} \& \bibinfo{author}{Senan, N.}
\newblock \bibinfo{title}{Violence video classification performance using deep neural networks}.
\newblock In \emph{\bibinfo{booktitle}{Recent Advances on Soft Computing and Data Mining: Proceedings of the Third International Conference on Soft Computing and Data Mining (SCDM 2018), Johor, Malaysia, February 06-07, 2018}}, \bibinfo{pages}{225--233} (\bibinfo{organization}{Springer}, \bibinfo{year}{2018}).

\bibitem{bagga2022violence}
\bibinfo{author}{Bagga, N.}, \bibinfo{author}{Singh, G.}, \bibinfo{author}{Balusamy, B.} \& \bibinfo{author}{Singh, A.~S.}
\newblock \bibinfo{title}{Violence detection in real life videos using convolutional neural network}.
\newblock In \emph{\bibinfo{booktitle}{2022 2nd International Conference on Advance Computing and Innovative Technologies in Engineering (ICACITE)}}, \bibinfo{pages}{872--876} (\bibinfo{organization}{IEEE}, \bibinfo{year}{2022}).

\bibitem{cheng2021rwf}
\bibinfo{author}{Cheng, M.}, \bibinfo{author}{Cai, K.} \& \bibinfo{author}{Li, M.}
\newblock \bibinfo{title}{Rwf-2000: an open large scale video database for violence detection}.
\newblock In \emph{\bibinfo{booktitle}{2020 25th International Conference on Pattern Recognition (ICPR)}}, \bibinfo{pages}{4183--4190} (\bibinfo{organization}{IEEE}, \bibinfo{year}{2021}).

\bibitem{aldahoul2021convolutional}
\bibinfo{author}{AlDahoul, N.} \emph{et~al.}
\newblock \bibinfo{title}{Convolutional neural network-long short term memory based iot node for violence detection}.
\newblock In \emph{\bibinfo{booktitle}{2021 IEEE International Conference on Artificial Intelligence in Engineering and Technology (IICAIET)}}, \bibinfo{pages}{1--6} (\bibinfo{organization}{IEEE}, \bibinfo{year}{2021}).

\bibitem{abdullah2023combination}
\bibinfo{author}{Abdullah, M. S. N.~B.}, \bibinfo{author}{Karim, H.~A.} \& \bibinfo{author}{AlDahoul, N.}
\newblock \bibinfo{title}{A combination of light pre-trained convolutional neural networks and long short-term memory for real-time violence detection in videos}.
\newblock \emph{\bibinfo{journal}{Methods}} \textbf{\bibinfo{volume}{400}}, \bibinfo{pages}{4000} (\bibinfo{year}{2023}).

\bibitem{wang2017review}
\bibinfo{author}{Wang, H.}, \bibinfo{author}{Yang, L.}, \bibinfo{author}{Wu, X.} \& \bibinfo{author}{He, J.}
\newblock \bibinfo{title}{A review of bloody violence in video classification}.
\newblock In \emph{\bibinfo{booktitle}{2017 International Conference on the Frontiers and Advances in Data Science (FADS)}}, \bibinfo{pages}{86--91} (\bibinfo{organization}{IEEE}, \bibinfo{year}{2017}).

\bibitem{doss2024comparative}
\bibinfo{author}{Doss, S.} \emph{et~al.}
\newblock \bibinfo{title}{Comparative analysis of news articles summarization using llms}.
\newblock In \emph{\bibinfo{booktitle}{2024 Asia Pacific Conference on Innovation in Technology (APCIT)}}, \bibinfo{pages}{1--6} (\bibinfo{organization}{IEEE}, \bibinfo{year}{2024}).

\bibitem{al2024analysis}
\bibinfo{author}{Al~Faraby, S.}, \bibinfo{author}{Romadhony, A.} \emph{et~al.}
\newblock \bibinfo{title}{Analysis of llms for educational question classification and generation}.
\newblock \emph{\bibinfo{journal}{Computers and Education: Artificial Intelligence}} \bibinfo{pages}{100298} (\bibinfo{year}{2024}).

\bibitem{nejjar2023llms}
\bibinfo{author}{Nejjar, M.}, \bibinfo{author}{Zacharias, L.}, \bibinfo{author}{Stiehle, F.} \& \bibinfo{author}{Weber, I.}
\newblock \bibinfo{title}{Llms for science: Usage for code generation and data analysis}.
\newblock \emph{\bibinfo{journal}{Journal of Software: Evolution and Process}} \bibinfo{pages}{e2723} (\bibinfo{year}{2023}).

\bibitem{huang2024generating}
\bibinfo{author}{Huang, C.-Y.}, \bibinfo{author}{Wei, J.} \& \bibinfo{author}{Huang, T.-H.}
\newblock \bibinfo{title}{Generating educational materials with different levels of readability using llms}.
\newblock \emph{\bibinfo{journal}{arXiv preprint arXiv:2406.12787}}  (\bibinfo{year}{2024}).

\bibitem{minaee2024large}
\bibinfo{author}{Minaee, S.} \emph{et~al.}
\newblock \bibinfo{title}{Large language models: A survey}.
\newblock \emph{\bibinfo{journal}{arXiv preprint arXiv:2402.06196}}  (\bibinfo{year}{2024}).

\bibitem{piot2024decoding}
\bibinfo{author}{Piot, P.} \& \bibinfo{author}{Parapar, J.}
\newblock \bibinfo{title}{Decoding hate: Exploring language models' reactions to hate speech}.
\newblock \emph{\bibinfo{journal}{arXiv preprint arXiv:2410.00775}}  (\bibinfo{year}{2024}).

\bibitem{yenduri2024gpt}
\bibinfo{author}{Yenduri, G.} \emph{et~al.}
\newblock \bibinfo{title}{Gpt (generative pre-trained transformer)--a comprehensive review on enabling technologies, potential applications, emerging challenges, and future directions}.
\newblock \emph{\bibinfo{journal}{IEEE Access}}  (\bibinfo{year}{2024}).

\bibitem{wu2023brief}
\bibinfo{author}{Wu, T.} \emph{et~al.}
\newblock \bibinfo{title}{A brief overview of chatgpt: The history, status quo and potential future development}.
\newblock \emph{\bibinfo{journal}{IEEE/CAA Journal of Automatica Sinica}} \textbf{\bibinfo{volume}{10}}, \bibinfo{pages}{1122--1136} (\bibinfo{year}{2023}).

\bibitem{deng2024vision}
\bibinfo{author}{Deng, J.}, \bibinfo{author}{Heybati, K.} \& \bibinfo{author}{Shammas-Toma, M.}
\newblock \bibinfo{title}{When vision meets reality: Exploring the clinical applicability of gpt-4 with vision} (\bibinfo{year}{2024}).

\bibitem{mu2024embodiedgpt}
\bibinfo{author}{Mu, Y.} \emph{et~al.}
\newblock \bibinfo{title}{Embodiedgpt: Vision-language pre-training via embodied chain of thought}.
\newblock \emph{\bibinfo{journal}{Advances in Neural Information Processing Systems}} \textbf{\bibinfo{volume}{36}} (\bibinfo{year}{2024}).

\bibitem{gupta2022towards}
\bibinfo{author}{Gupta, T.}, \bibinfo{author}{Kamath, A.}, \bibinfo{author}{Kembhavi, A.} \& \bibinfo{author}{Hoiem, D.}
\newblock \bibinfo{title}{Towards general purpose vision systems: An end-to-end task-agnostic vision-language architecture}.
\newblock In \emph{\bibinfo{booktitle}{Proceedings of the IEEE/CVF Conference on Computer Vision and Pattern Recognition}}, \bibinfo{pages}{16399--16409} (\bibinfo{year}{2022}).

\bibitem{antol2015vqa}
\bibinfo{author}{Antol, S.} \emph{et~al.}
\newblock \bibinfo{title}{Vqa: Visual question answering}.
\newblock In \emph{\bibinfo{booktitle}{Proceedings of the IEEE international conference on computer vision}}, \bibinfo{pages}{2425--2433} (\bibinfo{year}{2015}).

\bibitem{Gemini_15_technical_report}
\bibinfo{author}{Gemini~Team, G.}
\newblock \bibinfo{title}{Gemini 1.5 technical report}.
\newblock \bibinfo{howpublished}{\url{https://storage.googleapis.com/deepmind-media/gemini/gemini\_v1\_5\_report.pdf}} (\bibinfo{year}{2024}).

\bibitem{Introducing_Gemini_15}
\bibinfo{title}{Introducing gemini 1.5, google’s next-generation ai model}.
\newblock \bibinfo{howpublished}{\url{https://blog.google/technology/ai/google-gemini-next-generation-model-february-2024/\#architecture}} (\bibinfo{year}{2024}).

\bibitem{Hello_GPT-4o}
\bibinfo{title}{Hello gpt-4o}.
\newblock \bibinfo{howpublished}{\url{https://openai.com/index/hello-gpt-4o/ /}} (\bibinfo{year}{2024}).

\bibitem{GPT-4o}
\bibinfo{author}{OpenAI}.
\newblock \bibinfo{title}{Gpt-4o} (\bibinfo{year}{2024}).
\newblock \urlprefix\url{https://platform.openai.com/docs/models/gpt-4o}.

\bibitem{hate_speech_offensive}
\bibinfo{author}{Face, H.}
\newblock \bibinfo{title}{tdavidson/hate\_speech\_offensive}.
\newblock \urlprefix\url{https://huggingface.co/datasets/tdavidson/hate_speech_offensive/tree/main/data}.

\bibitem{Gender-Based_Violence}
\bibinfo{author}{Dutta, G.}
\newblock \bibinfo{title}{Gender-based violence tweet classification}.
\newblock \urlprefix\url{https://www.kaggle.com/datasets/gauravduttakiit/gender-based-violence-tweet-classification?select=Train.csv}.

\bibitem{Adult-content-dataset}
\bibinfo{author}{Face, H.}
\newblock \bibinfo{title}{valurank/adult-content-dataset}.
\newblock \bibinfo{howpublished}{\url{https://huggingface.co/datasets/valurank/Adult-content-dataset}}.

\bibitem{hou2024bridging}
\bibinfo{author}{Hou, Y.} \emph{et~al.}
\newblock \bibinfo{title}{Bridging language and items for retrieval and recommendation}.
\newblock \emph{\bibinfo{journal}{arXiv preprint arXiv:2403.03952}}  (\bibinfo{year}{2024}).

\bibitem{Amazon_REcognizer_Content_Moderation}
\bibinfo{title}{How amazon shopping uses amazon rekognition content moderation to review harmful images in product reviews}.
\newblock \bibinfo{howpublished}{\url{https://aws.amazon.com/blogs/machine-learning/how-amazon-shopping-uses-amazon-rekognition-content-moderation-to-review-harmful-images-in-product-reviews/}}.

\bibitem{Amazon_Transcribe_Toxicity_Detection}
\bibinfo{title}{Why amazon transcribe toxicity detection?}
\newblock \bibinfo{howpublished}{\url{https://aws.amazon.com/transcribe/toxicity-detection/}}.

\bibitem{soliman2019violence}
\bibinfo{author}{Soliman, M.~M.} \emph{et~al.}
\newblock \bibinfo{title}{Violence recognition from videos using deep learning techniques}.
\newblock In \emph{\bibinfo{booktitle}{2019 ninth international conference on intelligent computing and information systems (ICICIS)}}, \bibinfo{pages}{80--85} (\bibinfo{organization}{IEEE}, \bibinfo{year}{2019}).

\bibitem{Llama_Guard}
\bibinfo{author}{Face, H.}
\newblock \bibinfo{title}{meta-llama/llama-guard-3-8b}.
\newblock \bibinfo{howpublished}{\url{https://huggingface.co/meta-llama/Llama-Guard-3-8B}}.

\bibitem{Llama_Guard_vision}
\bibinfo{author}{Face, H.}
\newblock \bibinfo{title}{meta-llama/llama-guard-3-11b-vision}.
\newblock \bibinfo{howpublished}{\url{https://huggingface.co/meta-llama/Llama-Guard-3-11B-Vision}}.

\bibitem{Llama}
\bibinfo{author}{Face, H.}
\newblock \bibinfo{title}{meta-llama/llama-3.1-8b-instruct}.
\newblock \bibinfo{howpublished}{\url{https://huggingface.co/meta-llama/Llama-3.1-8B-Instruct}}.

\bibitem{Llama_vision}
\bibinfo{author}{Face, H.}
\newblock \bibinfo{title}{meta-llama/llama-3.2-11b-vision-instruct}.
\newblock \bibinfo{howpublished}{\url{https://huggingface.co/meta-llama/Llama-3.2-11B-Vision-Instruct}}.

\bibitem{Gemini_Vertex}
\bibinfo{author}{Cloud, G.}
\newblock \bibinfo{title}{Gemini, google’s most capable model, is now available on vertex ai}.
\newblock \bibinfo{howpublished}{\url{https://cloud.google.com/blog/products/ai-machine-learning/gemini-support-on-vertex-ai}} (\bibinfo{year}{2023}).

\bibitem{safety-settings}
\bibinfo{author}{Gemini~Team, G.}
\newblock \bibinfo{title}{safety-settings}.
\newblock \bibinfo{howpublished}{\url{https://ai.google.dev/gemini-api/docs/safety-settings}} (\bibinfo{year}{2024}).

\end{thebibliography}

\end{document}